\documentclass[runningheads]{llncs}

\usepackage{eccv}

\usepackage{eccvabbrv}
\usepackage{array}
\usepackage{multirow}
\usepackage{graphicx}
\usepackage{booktabs}
\usepackage{multirow}
\usepackage{makecell}
\usepackage{enumitem}
\usepackage{xspace}
\usepackage{amsmath}
\DeclareMathOperator*{\argmax}{arg\,max}
\usepackage{pifont}
\newcommand{\cmark}{\ding{51}}
\newcommand{\xmark}{\ding{55}}
\usepackage[table]{xcolor}
\definecolor{LightBlue}{HTML}{f2f2f2}
\usepackage{tcolorbox}
\usepackage{xcolor}
\usepackage{listings}

\usepackage[accsupp]{axessibility}  

\usepackage[
    colorlinks=true,
    linkcolor=black!75,
    citecolor=eccvblue,
    urlcolor=eccvblue,
    breaklinks=true
]{hyperref}

\usepackage{orcidlink}

\newcommand{\OurMethod}{HiVG\xspace}

\begin{document}

\title{
Hierarchical SVG Tokenization: \\ 
Learning Compact Visual Programs for Scalable Vector Graphics Modeling
}

\titlerunning{Hierarchical SVG Tokenization}

\author{
Ximing Xing\inst{1,2}$^{*}$,
Ziteng Xue\inst{2},
Zhenxi Li\inst{1},
Weicong Liang\inst{1},
Linqing Wang\inst{1},
Zhantao Yang\inst{1},
Tiankai Hang\inst{1},
Zijin Yin\inst{1},
Qinglin Lu\inst{1},
Chunyu Wang\inst{1}$^\dagger$$^{\ddagger}$,
Qian Yu\inst{2}$^{\ddagger}$
}

\authorrunning{X. Xing et al.}

\institute{
Tencent\inst{1},
Visual Computing Group, Beijing\inst{2} \\
$^\dagger$ Project Leader. $^{\ddagger}$ Corresponding author. \\
\url{https://hy-hivg.github.io/}
}

\maketitle



\begin{figure*}[th]
\vspace{-1.5em}
\centering
\includegraphics[width=1.0\textwidth]{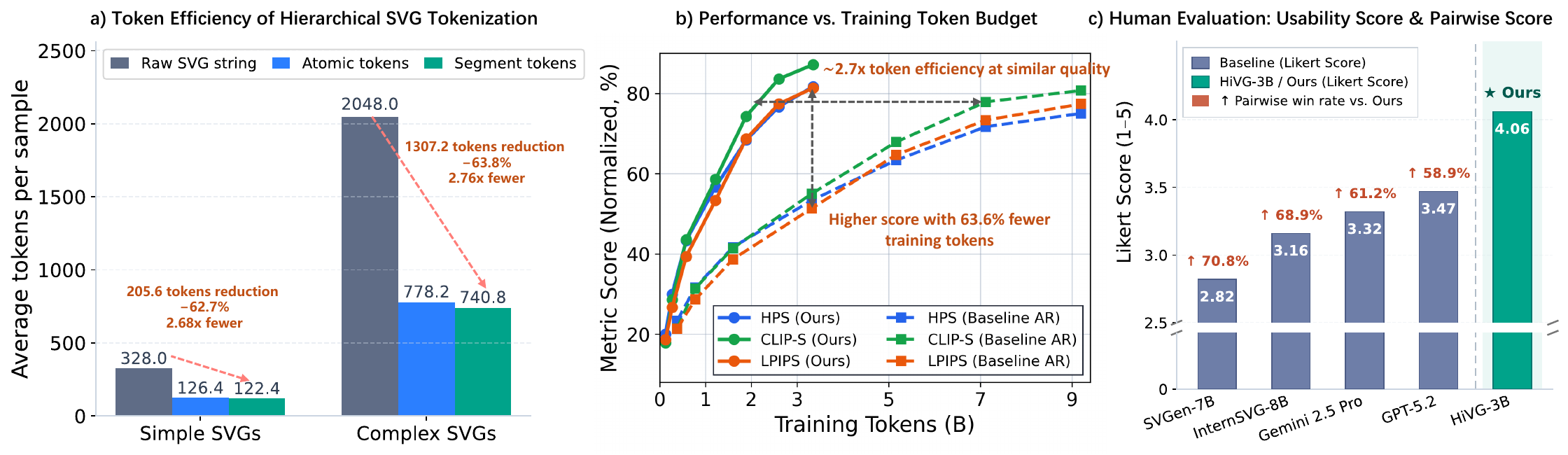}
\vspace{-2.3em}
\caption{
\textbf{Sequence-length compression, token-efficient scaling, and human evaluation.}
(a) HiVG tokenization compresses SVG sequences by 62.7\%--63.8\% (2.68$\times$--2.76$\times$).
(b) HiVG reaches comparable quality with approximately 2.7$\times$ fewer training tokens.
(c) HiVG achieves the best human evaluation results, scoring 4.06 in usability and winning 58.9\%--70.8\% in pairwise comparisons against baselines.
} \label{fig:teaser}
\vspace{-2.5em}
\end{figure*}
\begin{abstract}
Recent large language models have shifted SVG generation from differentiable rendering optimization to autoregressive program synthesis. 
However, existing approaches still rely on generic byte-level tokenization inherited from natural language processing, which poorly reflects the geometric structure of vector graphics. 
Numerical coordinates are fragmented into discrete symbols, destroying spatial relationships and introducing severe token redundancy, often leading to coordinate hallucination and inefficient long-sequence generation.
To address these challenges, we propose \OurMethod, a hierarchical SVG tokenization framework tailored for autoregressive vector graphics generation. 
\OurMethod decomposes raw SVG strings into structured \textit{atomic tokens} and further compresses executable command--parameter groups into geometry-constrained \textit{segment tokens}, substantially improving sequence efficiency while  preserving syntactic validity.
To further mitigate spatial mismatch, we introduce a Hierarchical Mean--Noise (HMN) initialization strategy that injects numerical ordering signals and semantic priors into new token embeddings. 
Combined with a curriculum training paradigm that progressively increases program complexity, \OurMethod enables more stable learning of executable SVG programs. 
Extensive experiments on both text-to-SVG and image-to-SVG tasks demonstrate improved generation fidelity, spatial consistency, and sequence efficiency compared with conventional tokenization schemes. Our code is publicly available at \url{https://github.com/ximinng/HiVG}.

\keywords{Autoregressive SVG Generation \and Hierarchical Representation \and Geometry-Aware Tokenization \and Structure-Preserving Token Compression}
\end{abstract}

\section{Introduction}
\label{sec:intro}

Scalable Vector Graphics (SVG) generation has recently attracted increasing attention due to its infinite-resolution rendering and highly compact representation.
Early methods~\cite{li2020differentiable,evolution_tian_2022,frans2022clipdraw,clipasso_vinker_2022,jain2023vectorfusion,xing2023diffsketcher,xing2024svgdreamer}, typically formulate SVG generation as a differentiable rendering or optimization problem, where vector primitives are iteratively adjusted to approximate a target image.
However, such approaches often suffer from high computational cost and limited ability to model the compositional structure of SVG programs.
With the rapid advancement of Large Language Models (LLMs), recent works have shifted towards treating SVG as executable code and generating it through autoregressive modeling~\cite{yang2025omnisvg,xing2025empowering,rodriguez2025starvector,wang2025internsvg,wu2025chat2svg}.


However, current LLM-based SVG generation methods still suffer from fundamental representation issues.
Alongside this architectural shift, we observe a concerning trend: existing methods inherit coordinate representations from the pre-trained LLM, which often leads to coordinate hallucination~\cite{yang2025omnisvg,huang2024opera}.
Recent works attempt to alleviate this through pre-processing of raw SVG strings (e.g., converting to relative coordinates~\cite{xing2025empowering} or flattened coordinates~\cite{yang2025omnisvg}).
Yet the fundamental problem remains unresolved: \textit{tokenized coordinates fail to reflect their underlying geometric relationships.}
Specifically, standard byte-level tokenizers~\cite{bpe_sennrich_2016} treat numerical coordinates as discrete strings rather than continuous spatial values (e.g., ``100'' is tokenized as ``1'', ``0'', ``0''). Consequently, this fragmentation not only destroys the inherent spatial relationships of the coordinates but introduces severe token redundancy during generation.

Beyond the coordinate hallucination, another key challenge lies in the inefficient representation of SVG sequences.
Existing works also struggle when generating complex SVG content~\cite{wang2025svgen,yang2025omnisvg}.
A common workaround is to expand the model context window.
However, we argue that this limitation fundamentally stems from the inherently low information density of raw SVG tokens compared to natural language.
While a semantic word usually requires only 1–2 tokens, even a simple SVG shape may be represented by a long string of drawing commands and coordinates, which may expand to tens or even hundreds of tokens after tokenization.
Such redundancy contradicts the structural compactness that makes SVG appealing in the first place. Above observations lead us to ask: \textbf{how can we rethink the tokenization paradigm to natively align with the underlying properties of vector graphics?}
\begin{figure*}[t]
\includegraphics[width=\textwidth]{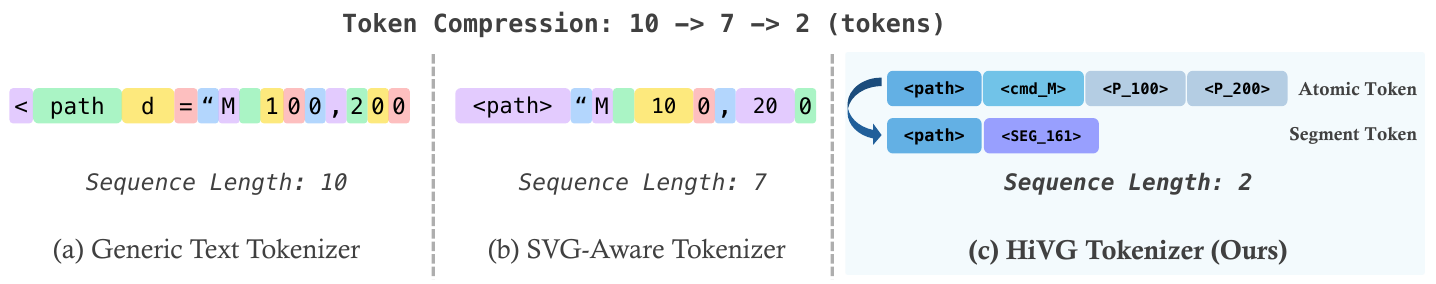}
\vspace{-2em}
\caption{
\textbf{Comparison of SVG tokenization strategies.}
(a) A generic LLM tokenizer~\cite{qwen2.5_2024,qwen2.5vl_2025} treats SVG as plain text and splits it into subword tokens, producing long token sequences.
(b) An SVG-aware tokenizer~\cite{xing2025empowering,wang2025internsvg} improves structural awareness by tokenizing SVG elements and attributes, but geometric primitives remain fragmented into many numeric coordinate tokens.
(c) Our HiVG tokenizer introduces a hierarchical representation that groups drawing commands together with their associated coordinates into reusable segment tokens, enabling substantial sequence compression (10 → 7 → 2).
}
\label{fig:tokenizer_compare}
\vspace{-2em}
\end{figure*}
\textbf{The above cues reveal that the devil is in the token compression}. 
We provide an affirmative answer to this challenge by proposing \textbf{HiVG}, a novel, \textbf{hi}erarchical S\textbf{VG} tokenizer tailored exclusively for autoregressive generation that decomposes vector graphics into structure-preserving components instead of flat byte streams.
As shown in Fig.~\ref{fig:tokenizer_compare}, we first transform raw SVG strings into foundational \textbf{atomic tokens} that strictly separate structures, drawing commands, coordinates, and attributes. 
To fully exploit the inherently renderable patterns of SVG, we design a merging strategy that compresses vector sequences into composite \textbf{segment tokens} under geometric constraints.
This hierarchical compression drastically shortens coordinate-heavy sequences while guaranteeing that every merged token remains a syntactically valid, executable geometric primitive. 
Under this hierarchical representation, segment tokens reduce sequence length by up to 63.8\% relative to raw-string tokenization on Qwen~\cite{qwen2.5vl_2025} (see Fig.~\ref{fig:teaser} (a)).

To further resolve the spatial mismatch caused by discretized coordinate tokens, we introduce a Hierarchical Mean-Noise (HMN) initialization strategy.
Instead of randomly initializing the newly introduced SVG vocabulary, HMN explicitly injects numeric ordering signals and semantic priors into token embeddings.
These signals are projected through a Gaussian–polynomial basis, enabling embeddings to preserve continuous spatial relationships among coordinates.
Our experiments demonstrate that such mathematically grounded initialization provides the model with native spatial awareness from the very beginning of training.
Combined with the hierarchical representation, HiVG reaches higher visual quality with approximately 2.7$\times$ fewer training tokens (see Fig.~\ref{fig:teaser} (b)).

In summary, our contributions are three-fold:
(1) We propose \OurMethod, a hierarchical SVG tokenization framework that decomposes raw SVG code into atomic tokens and compresses command--parameter groups into executable segment tokens, substantially reducing sequence length while preserving syntactic validity.
(2) We introduce Hierarchical Mean--Noise (HMN) initialization, which injects numeric ordering signals and semantic priors into new token embeddings to improve spatial awareness and coordinate consistency.
(3) We adopt a three-stage curriculum that progressively increases program depth, leading to more stable optimization and improved generalization to long SVG sequences on both text-to-SVG and image-to-SVG tasks.

\section{Related Work}
\subsection{Parametric Vector Graphics Paradigm}
Recent advancements in Scalable Vector Graphics (SVG) generation can be broadly distinguished by how they represent the underlying graphic data. 
Early optimization-based methods~\cite{li2020differentiable,evolution_tian_2022,frans2022clipdraw,clipasso_vinker_2022,jain2023vectorfusion,xing2023diffsketcher,xing2024svgdreamer} treat SVGs as collections of stroke parameters and iteratively optimize them via differentiable renderers. 
Another line of research projects SVG commands and their numerical attributes into continuous latent spaces to learn compact implicit representations~\cite{carlier2020deepsvg,strokenuwa_tang_2024,xing2024svgfusion}. 
More recently, research has shifted toward representing SVGs as sequences of discrete tokens. Consistent with this trend, recent efforts closely mirror the evolution of broader LLM frameworks, incorporating large-scale datasets~\cite{rodriguez2025starvector,wang2025internsvg,yang2025omnisvg,xing2025empowering,wang2025svgen}, reinforcement learning techniques~\cite{xing2025reason,rodriguez2025rendering,wang2025svgen}, and unified tasks~\cite{yang2025omnisvg,li2025unisvg,wang2025internsvg}. 
While these high-level integrations have led to notable progress, tokenization itself remains relatively underexplored. Although some works embed SVG commands to better capture their semantics~\cite{xing2025empowering,yang2025omnisvg}, this does not necessarily yield a principled tokenization scheme. This observation motivates our work.


\subsection{Token Representation and Compression}
Efficient token representation is fundamental to autoregressive sequence modeling.
A longstanding belief holds that compression is closely connected to
intelligence, with some researchers suggesting that they are fundamentally
equivalent~\cite{huang2024compression,deletang2023language}. In the field of language modeling, this principle is exemplified by Byte-Pair Encoding (BPE)~\cite{bpe_sennrich_2016,kudo2018sentencepiece}, which effectively mitigates token sparsity by merging frequently co-occurring characters into robust subword representations. Building upon this, a growing body of works across diverse modalities has explored task-specific compression strategies to adapt complex data for autoregressive modeling. For example, FAST~\cite{pertsch2025fast} compresses continuous robot action chunks to learn generalizable robotic behaviors, while FreeMesh~\cite{liu2025freemesh} quantifies 3D mesh sequence learnability by balancing entropy and compression. In the Computer-Aided Design (CAD) domain, CAD-GPT~\cite{wang2025cadgpt} compresses 3D spatial parameters and 2D sketch coordinates into a 1D linguistic token space to enhance the spatial reasoning capabilities. 
In the domain of SVG, prior methods have also explored various tokenization schemes.
For example, DeepSVG~\cite{carlier2020deepsvg} represents SVG paths as sequences of drawing commands with associated parameters, while LLM4SVG~\cite{xing2025empowering} serializes SVG elements into textual command tokens for autoregressive generation. 
However, these approaches typically operate at the level of individual coordinates or drawing commands, resulting in lengthy token sequences that hinder both training and inference efficiency.
Different from the free-form combination or coordinate-level discretization seen in these prior works, we identify renderable units under geometric constraints to achieve structural token compression for SVG generation. This structural compression yields substantially shorter sequences while preserving geometric fidelity, leading to a superior compression rate and significantly improved computational efficiency.


\begin{figure*}[!t]
\includegraphics[width=\textwidth]{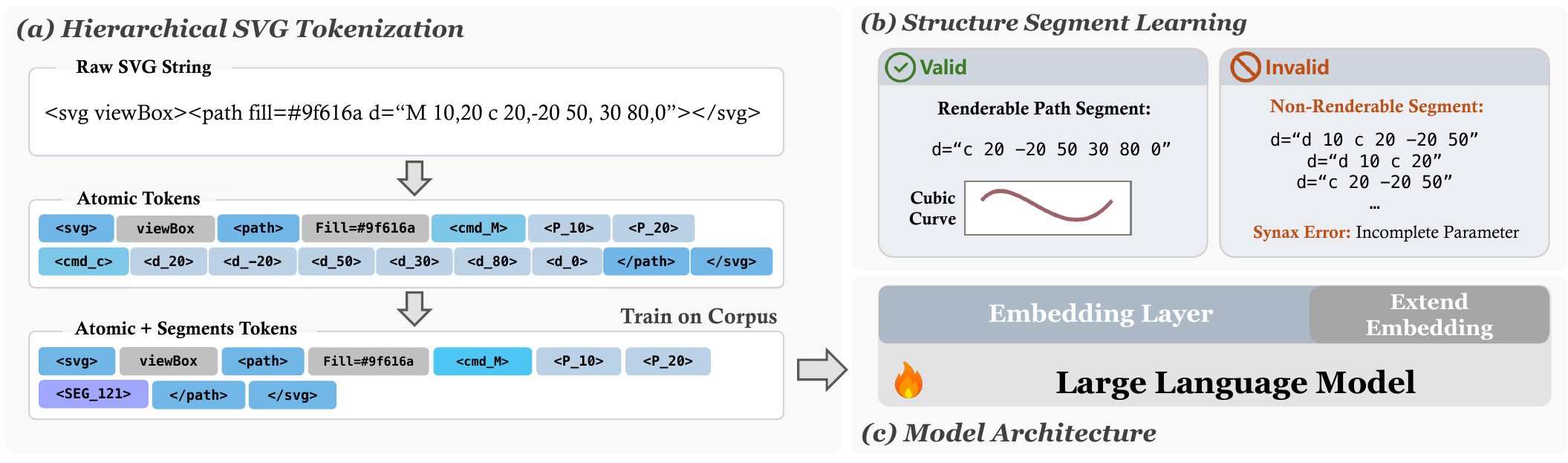}
\vspace{-2em}
\caption{
\textbf{Overview of HiVG.}
(a) \emph{Hierarchical SVG Tokenization} (\cref{sec:hierarchical_svg_tokenization}).
Raw SVG strings are first decomposed into atomic tokens and further compressed into executable segment tokens via structure-aware merging.
Only complete command–parameter units are merged to ensure geometric validity.
(b) \emph{Structure Segment Learning}.
Segment tokens are learned from a large SVG corpus by discovering renderable command–coordinate groups while discarding merges that violate syntactic or geometric constraints.
(c) \emph{Model Architecture}.
Atomic and Segment tokens extend the embedding space of the base LLM, while training progressively increases program depth to stabilize structural abstraction and global composition.
}
\label{fig:method}
\vspace{-1em}
\end{figure*}
\section{HiVG: Hierarchical SVG Modeling}
\label{sec:method}

\subsection{Hierarchical SVG Tokenization}
\label{sec:hierarchical_svg_tokenization}
A key challenge in autoregressive SVG generation arises from the program-like nature of vector graphics.
Although a typical icon contains only a small number of visual primitives, its serialized SVG representation is dominated by long sequences of numeric coordinates.
Consequently, low-level coordinate tokens overwhelm the context, while higher-level structural signals become sparse.
This imbalance makes it difficult for language models to infer element boundaries, preserve structural validity, and reason about geometric relationships across distant parts of the sequence.

To address this issue, we introduce a hierarchical tokenization scheme that decomposes SVG programs into structured atomic tokens and further compresses coordinate-heavy command segments into reusable geometric primitives.
An overview of the proposed HiVG framework is illustrated in Fig.~\ref{fig:method}.

\noindent\textbf{Atomic SVG Tokens.}
We first transform raw SVG strings into a sequence of atomic tokens that preserve full rendering executability.
The atomic vocabulary is partitioned into four disjoint categories:
\begin{equation}
\small
\mathcal{V}_\text{atomic}
=
\mathcal{V}_\text{struct}
\;\cup\;
\mathcal{V}_\text{cmd}
\;\cup\;
\mathcal{V}_\text{coord}
\;\cup\;
\mathcal{V}_\text{attr}.
\label{eq:atomic_tok}
\end{equation}
\noindent Here, $\mathcal{V}_\text{struct}$ contains structure tokens that define SVG elements and hierarchical layout (e.g., \texttt{<svg>}, \texttt{<path>});
$\mathcal{V}_\text{cmd}$ consists of path operators such as \texttt{<cmd\_M>} and \texttt{<cmd\_C>};
$\mathcal{V}_\text{attr}$ represents visual attributes including color and opacity.
The coordinate vocabulary $\mathcal{V}_\text{coord}$ encodes geometric positions.
Given a canvas of size $(W,H)$, raw coordinates are first normalized to the canvas range and uniformly quantized into discrete integer bins, each mapped to a coordinate token.

To improve compositional regularity, path parameters are represented primarily using relative coordinates.
Specifically, the first command in each path uses absolute coordinates to establish the starting position, while subsequent parameters are expressed relative to the previous point.
This representation reduces global translation variance and exposes recurring geometric patterns across SVG programs.
As a result, relative coordinates tend to increase the frequency of repeated command–coordinate groups in the corpus, which facilitates the discovery of reusable geometric primitives during segment learning.

Finally, each SVG command has a fixed parameter arity defined by the SVG specification (e.g., \texttt{lineto} requires two coordinates, while cubic B\'{e}zier curves require six).
This constraint naturally defines executable command–parameter groups consisting of a drawing operator and its required coordinates.
We refer to such units as \emph{segments}, which form the basic geometric primitives used for higher-level token construction.
Figure~\ref{fig:tokenizer_compare} illustrates how grouping commands with their parameters enables compact segment-level representations.

\begin{figure*}[t]
\includegraphics[width=\textwidth]{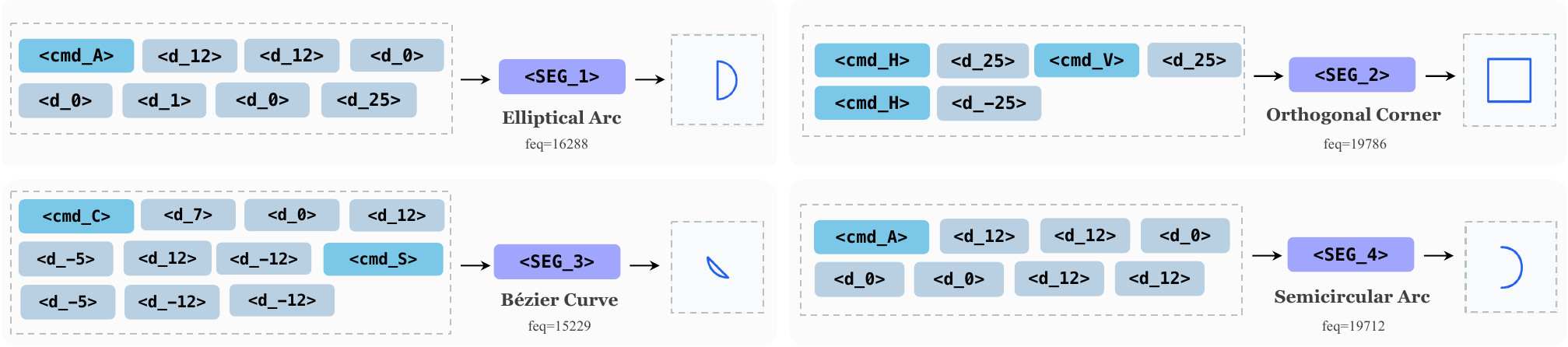}
\vspace{-2em}
\caption{\textbf{Learned segment tokens as renderable geometric primitives}. Segment-level merging preserves syntactic validity and geometric coherence while shortening token sequences and improving efficiency.
}
\label{fig:segment_token}
\vspace{-1em}
\end{figure*}

\noindent\textbf{Segment Tokens via Structure Segment Learning.} 
As shown in Fig.~\ref{fig:tokenizer_compare}(a,b), conventional tokenization strategies either treat SVG code as plain text or tokenize elements and attributes independently.
In both cases, geometric primitives are fragmented into long sequences of coordinate tokens, leading to inefficient and structurally fragmented representations.

To address this issue, we perform token merging over segments rather than individual tokens.
Formally, a segment is defined as a command token together with all of its coordinate parameters:
\begin{equation}
\small
s = (\texttt{<cmd>}, c_1, \ldots, c_k),
\label{eq:segment}
\end{equation}
\noindent where $k$ is uniquely determined by the command type.
Let $\mathcal{S} = \{s_1, s_2, \ldots\}$ denote the multiset of segments extracted from atomic token sequences.
We then perform iterative pair merging over $\mathcal{S}$.
At iteration $t$, the most frequent adjacent segment pair is selected
\begin{equation}
\small
(s_i^*, s_j^*) = \argmax_{(s_i, s_j)} \mathrm{count}(s_i, s_j),
\label{eq:bpe_merge}
\end{equation}
\noindent and replaced with a new composite segment token if its frequency exceeds $f_{\min}$.
After $M$ merging iterations, we obtain a vocabulary of learned segment tokens representing frequently occurring geometric primitives.

Importantly, merging is restricted to segment boundaries, while structure and attribute tokens remain unchanged.
As a result, all learned tokens correspond to renderable segment groups as shown in Fig.~\ref{fig:segment_token}, ensuring syntactic validity and geometric coherence.
This segment-level representation significantly reduces sequence length and improves token efficiency.

\begin{figure*}[t]
\includegraphics[width=\textwidth]{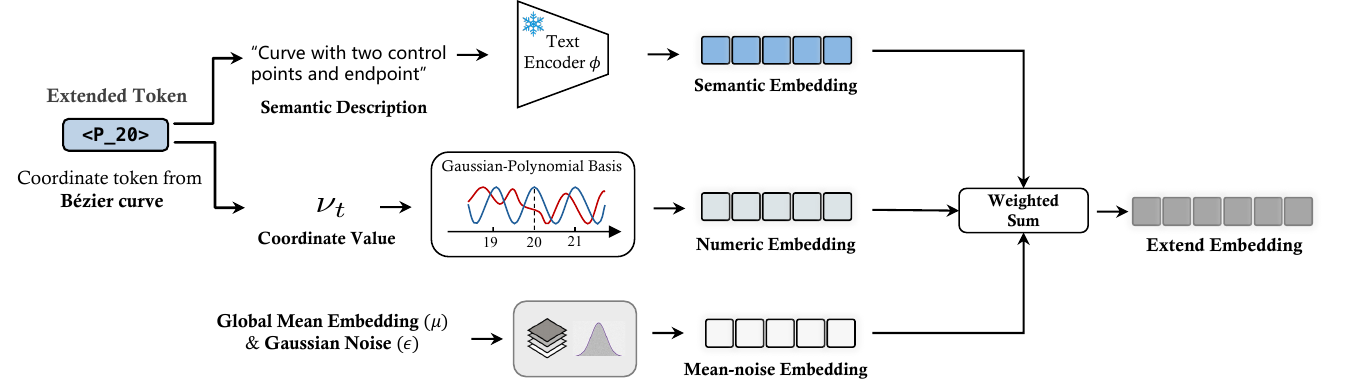}
\vspace{-2em}
\caption{\textbf{HMN initialization for structured SVG tokens}. Each new token is initialized by combining a global mean-noise prior with a semantic embedding from its textual description; for coordinate tokens, an additional numeric embedding derived from the normalized value through Gaussian–Polynomial basis encoding is added, while non-numeric tokens omit the numeric branch.
}
\label{fig:token_init}
\vspace{-1em}
\end{figure*}

\subsection{Token Initialization Strategy}
\label{sec:token_init}
Extending the vocabulary of a pretrained language model with domain-specific tokens requires careful embedding initialization.
A common practice initializes new embeddings either from isotropic Gaussian noise or from the global mean of the pretrained vocabulary.
However, such strategies ignore the internal structure of newly introduced tokens.

For structured vocabularies such as SVG primitives, tokens encode heterogeneous semantics, including element categories, geometric operators, and ordered numeric coordinates. 
We therefore introduce \textbf{Hierarchical Mean–Noise (HMN)} initialization, which combines semantic priors with a structured numeric perturbation. The detailed initialization is illustrated in Fig.~\ref{fig:token_init}.

For each newly added token $t$, its embedding is initialized as
\begin{equation}
\small
\mathbf{e}_t
=
\lambda_\mu \boldsymbol{\mu}
+
\lambda_n \boldsymbol{\epsilon}
+
w_{\mathrm{sem}}\, \phi(\mathrm{desc}_t)
+
w_{\mathrm{num}}\, \mathbf{d}_t,
\label{eq:hmn}
\end{equation}
\noindent where $\boldsymbol{\mu}$ denotes the mean embedding of the original vocabulary $\mathcal{V}_0$, 
$\boldsymbol{\epsilon} \sim \mathcal{N}(\mathbf{0}, \sigma^2\mathbf{I})$ introduces stochastic perturbation, and $\phi(\cdot)$ maps the textual description of token $t$ into the pretrained embedding space using frozen model weights.
The final term $\mathbf{d}_t$ encodes numeric information for coordinate tokens.

To construct $\mathbf{d}_t$, the scalar coordinate value $v_t$ (normalized to $[0,1]$) is first encoded using a low-dimensional basis representation combining Gaussian radial basis functions~\cite{randomfeature2007rahimi} and polynomial features.
This encoding captures both local smoothness and global ordering among coordinate values.
The resulting representation is then projected to the model embedding dimension using a fixed random projection matrix inspired by the Johnson–Lindenstrauss transform~\cite{ghojogh2021johnson}.
Finally, the projected vector is normalized to unit length and used as a small directional perturbation.

This design allows semantic information to remain dominant in the embedding space while numeric structure provides a consistent directional bias for coordinate tokens.
Consequently, HMN preserves distributional alignment with the pretrained vocabulary while injecting structured geometric information during initialization.

\subsection{Curriculum Training Paradigm}
\label{sec:curriculum_training}
Autoregressive SVG generation requires simultaneously aligning newly introduced structured tokens with the pretrained embedding space and modeling long-range geometric dependencies.
Direct training over the full sequence spectrum often destabilizes optimization.
We therefore adopt a structure-aware curriculum that progressively increases effective program depth.

\noindent\textbf{Stage 1: Embedding Alignment.}
Training begins with atomic SVG tokens and moderate-length sequences.
This stage aligns newly introduced tokens with the pretrained embedding manifold while stabilizing local geometric transitions.

\noindent\textbf{Stage 2: Structural Abstraction.}
Segment tokens are then activated, shifting learning from primitive transitions to executable geometric units.
The dependency horizon expands while preserving token-space stability.

\noindent\textbf{Stage 3: Global Composition.}
Finally, full-length SVG programs are introduced.
The model focuses on layout coherence and long-range inter-path dependencies.

Each stage expands the training distribution without discarding earlier regimes, separating embedding alignment, structural abstraction, and global composition into distinct optimization phases.

\section{Experiments}
\label{sec:experiments}

\subsection{Experimental Setup}
\label{sec:exp_setting}

\noindent\textbf{Dataset Construction.}
We construct our training corpus by merging three open-source SVG datasets and performing cross-source deduplication, resulting in 2.45M unique SVG samples covering diverse vector graphic categories, including icons, emojis, logos, and interface elements. Before tokenization, we apply a unified filtering and preprocessing pipeline to improve rendering consistency and representation quality. In particular, we remove malformed, unsafe, and non-renderable content, normalize SVG structure and styling, resolve reusable elements and transformations into explicit geometry, map all samples into a unified coordinate space, and quantize coordinates into a discrete tokenizer-friendly format. Samples that remain unstable after preprocessing are discarded. 
More detailed descriptions of dataset construction, filtering, and preprocessing are provided in Sec.~\ref{supp:dataset} of the supplementary material.

\noindent\textbf{Training Details.}
We fully fine-tune Qwen2.5-VL-3B-Instruct~\cite{qwen2.5vl_2025} under a supervised fine-tuning (SFT) setting.
The vision tower and multi-modal projector are frozen, while the language model and newly introduced SVG token embeddings are optimized.
All experiments are conducted at a fixed canvas resolution of $784 \times 784$.
We train for 2 epochs using AdamW with a learning rate of $1\times10^{-5}$ and a warmup ratio of 0.2.
New SVG tokens are initialized using the proposed Hierarchical Mean-Noise strategy.
The three-stage curriculum is implemented by progressively expanding the training dataset with increasing sequence length ranges while keeping optimization hyperparameters fixed.
More detailed descriptions of hyperparameters (\cref{supp:hyperparameters}) and training prompt template (\cref{supp:prompt_template}) are provided in Sec.~\ref{supp:implementation} of the supplementary material.

\noindent\textbf{SVG Token Vocabulary.}
At a fixed canvas resolution of $784{\times}784$, our atomic SVG vocabulary contains 2,450 tokens.
It consists of 2,384 coordinate tokens and 66 non-coordinate tokens.
The coordinate set includes 795 absolute position tokens ($\texttt{P\_0} \sim \texttt{P\_794}$) and 1,589 relative offset tokens ($\texttt{d\_-794} \sim \texttt{d\_794}$), enabling both absolute anchoring and full-range relative moves.
The non-coordinate set includes 42 structure tokens (21 SVG elements with paired open/close tags), 20 path-command tokens (10 commands with absolute/relative variants), and 4 arc-flag tokens (\texttt{large\_\{0,1\}}, \texttt{sweep\_\{0,1\}}).
Segment tokens are learned on top of this atomic vocabulary via Structure Segment Learning.

\noindent\textbf{Evaluation.}
We evaluate structural validity, semantic alignment, visual fidelity, diversity, and perceptual quality under both text-to-SVG and image-to-SVG settings.
\textit{(1) Validity and Efficiency.}
We report render success rate (Render), average token count (TokCnt), path count (PathCnt), and path command count (CmdCnt).
Lower TokCnt, PathCnt, and CmdCnt indicate more compact SVG programs while maintaining rendering fidelity.
\textit{(2) Semantic and Visual Quality.}
For text-to-SVG, we measure CLIP~\cite{clip_Radford_2021} similarity between rendered images and text prompts.
For image-to-SVG, we additionally report CLIP-visual similarity (CLIP-S) between the rendered image and the input reference image, as well as SSIM and LPIPS to assess structural fidelity and perceptual similarity.
\textit{(3) Diversity and Preference.}
To quantify sample diversity, we extract DINOv2-ViT-Large~\cite{dinov2_oquab_2024} features from generated images and compute the average pairwise cosine similarity across $L$ samples.
Diversity is defined as
\begin{equation}
\small
\mathrm{Diversity} =
1 - \frac{2}{L(L-1)} \sum_{i<j} \cos\!\left(x_\theta^{(i)}, x_\theta^{(j)}\right),
\end{equation}
where $x_\theta^{(i)}$ denotes the DINO feature of the $i$-th sample.
Higher diversity corresponds to lower feature similarity among generated outputs.

Perceptual quality is further evaluated using HPSv2~\cite{hpsv2_Wu_2023}, ImageReward~\cite{xu2023imagereward}, PickScore (PickS)~\cite{kirstain2023pickApic}, and Aesthetic score (Aes)~\cite{aesthetic_christoph_2022}.

\begin{figure*}[t]
\includegraphics[width=\textwidth]{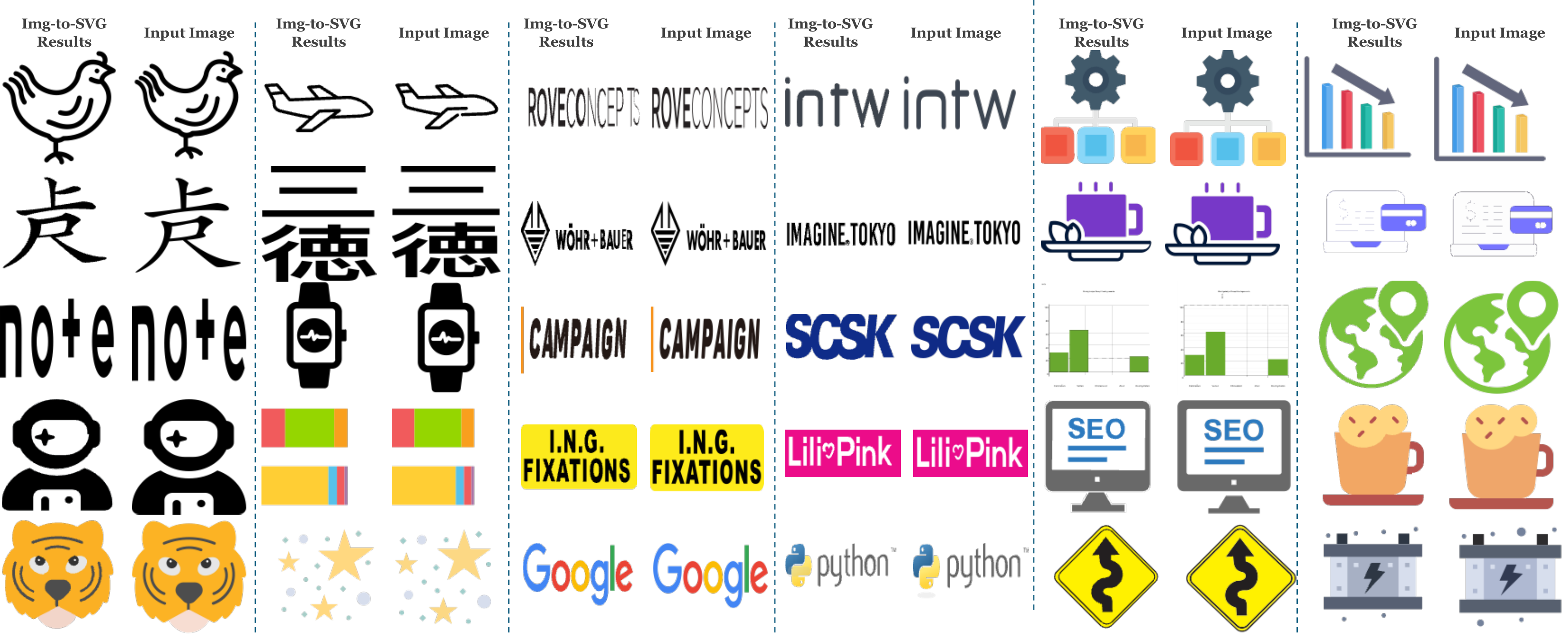}
\vspace{-2em}
\caption{
\textbf{Image-to-SVG generation results.}
For each example, the raster input image is shown on the right and the generated SVG rendering on the left.
The examples include icons, logos, typography, UI elements, and emoji-style graphics.
} \label{fig:ours_img2svg}
\vspace{2em} 
\vspace{-1em}
\includegraphics[width=\textwidth]{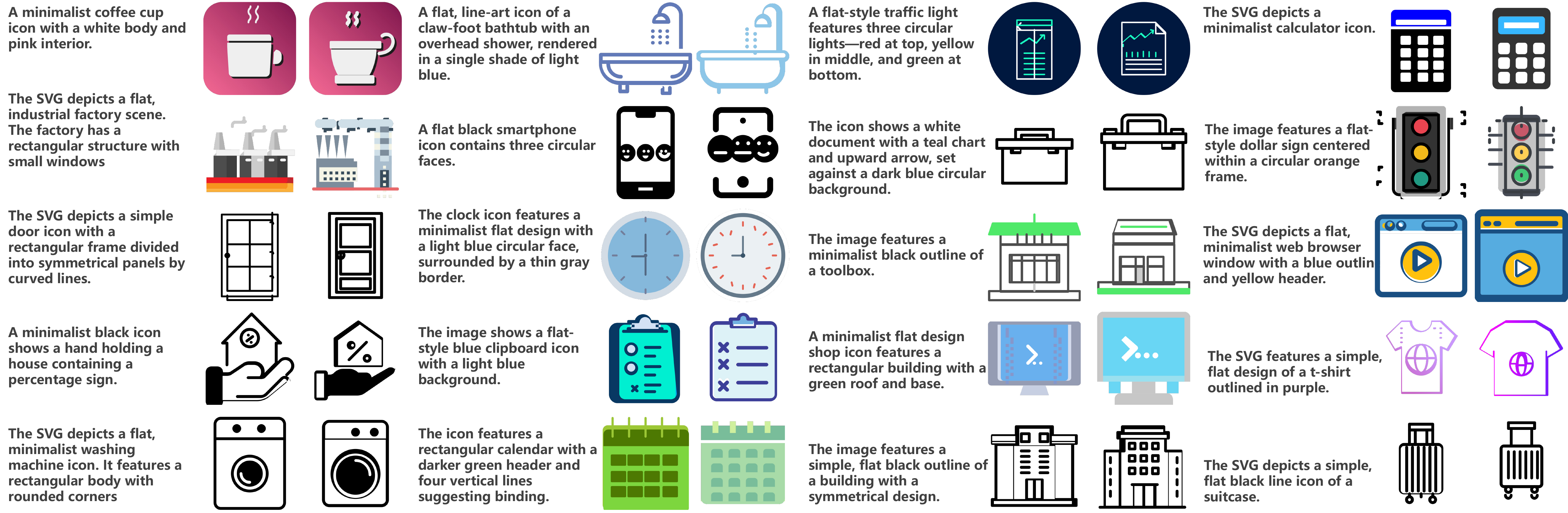}
\vspace{-2em}
\caption{
\textbf{Text-to-SVG generation results.}
Text prompts are shown on the left and the generated SVG renderings on the right.
The examples cover various object types such as household items, UI elements, buildings, clothing, and symbolic icons.
}\label{fig:ours_text2svg}
\vspace{-1em}
\end{figure*}
\begin{figure*}[!htp]
\centering
\includegraphics[width=1.0\textwidth]{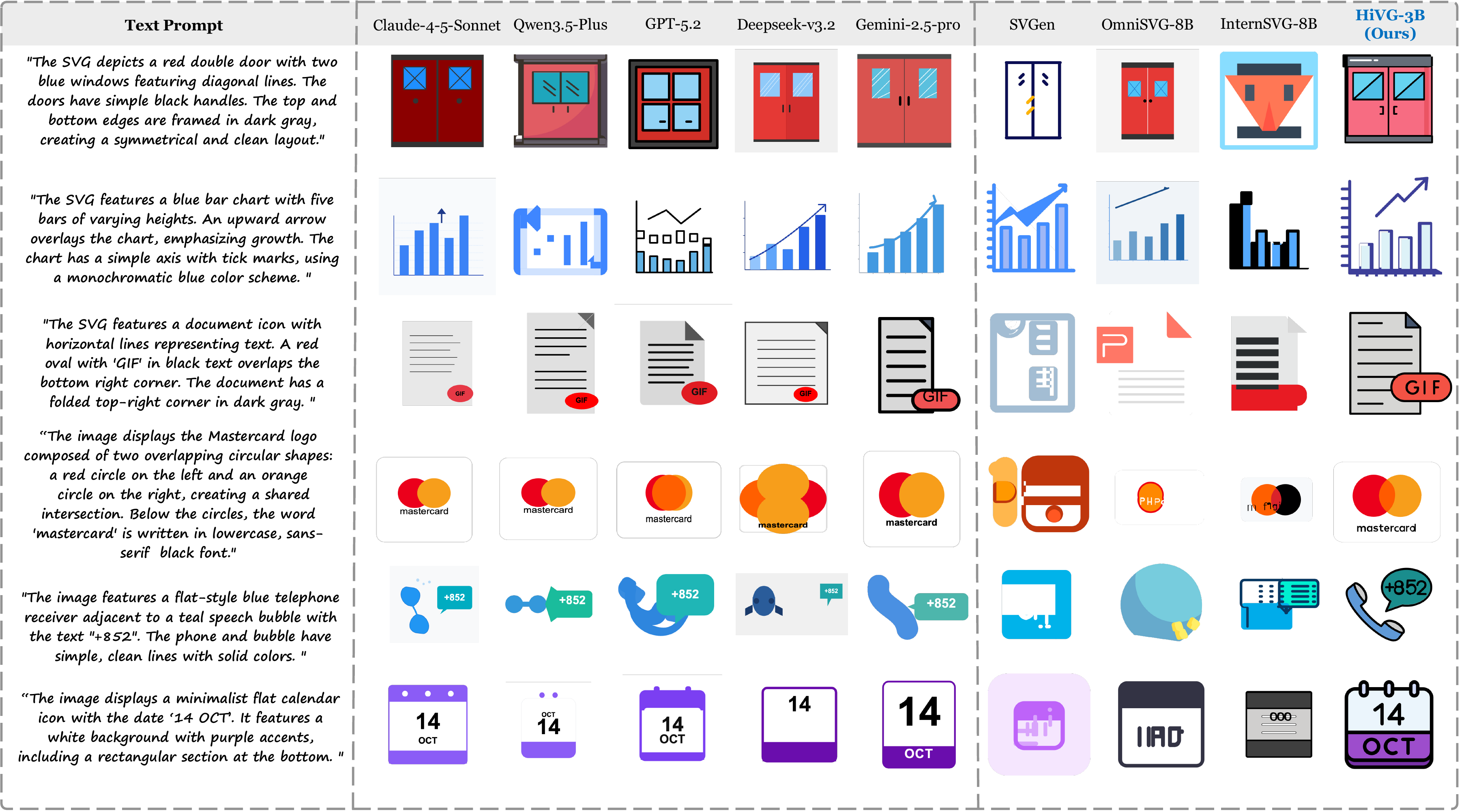} 
\vspace{-2em}
\caption{\textbf{Text-to-SVG comparison.} 
Each row corresponds to a text prompt (left). Columns show SVG renderings generated by different methods. Compared with existing models, HiVG produces SVG outputs with more consistent layout structure and better alignment with the prompt description.
}
\label{fig:text2svg_comparison}
\vspace{2em} 
\vspace{-1em}
\centering
\includegraphics[width=1.0\textwidth]{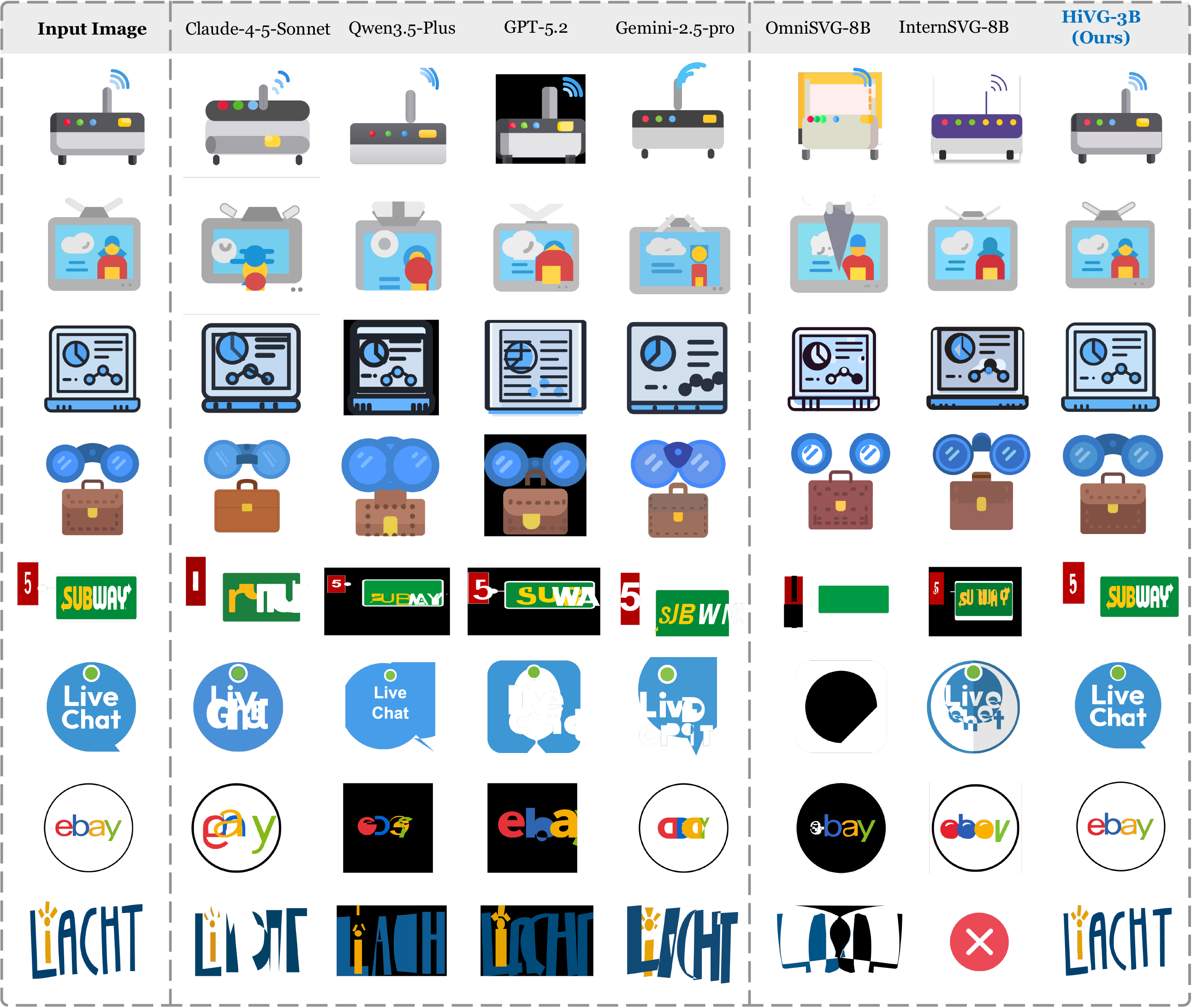}
\vspace{-2em}
\caption{\textbf{Image-to-SVG comparison.}
The first column shows the raster input image, and the remaining columns show SVG reconstructions generated by different methods.
}
\label{fig:img2svg_comparison}
\vspace{-1em}
\end{figure*}

\subsection{Qualitative and Quantitative Analysis}
\label{sec:quality_and_quantity}
\noindent\textbf{Quantitative Results.}
Table~\ref{tab:main_results} reports the quantitative comparison on both Image-to-SVG and Text-to-SVG tasks.
Compared with existing SVG generation models, HiVG achieves competitive or superior performance across multiple metrics.
This improvement suggests that grouping commands with their parameters
reduces long-range dependencies and improves structural consistency.
Notably, improvements in CLIP-S and LPIPS indicate that segment-level tokens better preserve global geometry while reducing local coordinate drift.
On Image-to-SVG reconstruction, our method obtains strong CLIP-S and aesthetic scores while maintaining stable structural validity and visual similarity.
On Text-to-SVG generation, HiVG achieves higher PickScore and competitive CLIP and HPS scores, indicating improved prompt alignment and perceptual quality.

\noindent\textbf{Qualitative Results.}
Figures~\ref{fig:ours_img2svg},~\ref{fig:ours_text2svg} show representative outputs generated by HiVG.
For Image-to-SVG reconstruction, the model accurately preserves object shapes, typography, and layout structures across icons, logos, and UI-style graphics.
For Text-to-SVG generation, HiVG produces visually coherent SVG outputs that follow the prompt description while maintaining geometric layouts.

\noindent\textbf{Comparison with Existing Methods.}
Figures~\ref{fig:text2svg_comparison} and~\ref{fig:img2svg_comparison} provide side-by-side comparisons with recent large multimodal models and SVG generation approaches.
In Text-to-SVG generation, several baselines produce incomplete shapes, incorrect layouts, or text mismatches, while HiVG generates more structurally consistent SVG programs.
In Image-to-SVG reconstruction, competing methods often introduce geometric distortions or color inconsistencies, whereas HiVG better preserves the global structure and visual details of the input image. It is worth noting that HiVG excels not only at generating iconographic elements, but also at producing textual content with remarkable consistency, a capability rarely achieved by existing methods.

\begin{table*}[!tp]
\centering
\caption{\textbf{Quantitative comparison on Img2SVG and Text2SVG tasks.} `-' denotes the method can only accept text input.}
\label{tab:main_results}
\vspace{-1em}
\setlength{\tabcolsep}{4pt}
\resizebox{\textwidth}{!}{
\begin{tabular}{l|cccccc|cccc}
\toprule
\multicolumn{1}{c|}{\multirow{2}{*}[-0.6ex]{\textbf{Method}}}& \multicolumn{6}{c|}{\textbf{Img2SVG}}
& \multicolumn{4}{c}{\textbf{Text2SVG}} \\
\cmidrule(r){2-7} \cmidrule(l){8-11}
& SSIM$\uparrow$
& LPIPS$\downarrow$
& CLIP-S$\uparrow$
& PickS$\uparrow$
& HPS$\uparrow$
& Aes$\uparrow$
& CLIP$\uparrow$
& PickS$\uparrow$
& HPS$\uparrow$
& Aes$\uparrow$ \\
\midrule
DeepSeekv3.2~\cite{deepseekai2025deepseekv32}
& - & - & - & - & - & -
& 0.272 & 20.331 & 0.192 & 4.594 \\
Qwen3.5 Plus~\cite{qwen3_5_2026}
& 0.775 & 0.228 & 0.896 & 22.019 & 0.175 & 4.672
& 0.291 & 20.972 & 0.206 & 4.671 \\
Gemini-2.5-pro~\cite{google2025gemini}
& 0.790 & 0.215 & 0.904 & 22.346 & 0.185 & 4.732
& 0.284 & 20.943 & 0.210 & 4.765 \\
GPT-5.2~\cite{openai2025gpt5}
& 0.780 & 0.205 & 0.930 & 23.977 & 0.222 & 4.841
& 0.291 & 21.268 & 0.214 & 4.806 \\
Claude-Sonnet-4.5~\cite{claude45sonnet_modelcard_2025}
& 0.669 & 0.292 & 0.842 & 22.012 & 0.164 & 4.435
& 0.281 & 20.562 & 0.195 & 4.711 \\
\midrule
SVGen-7B~\cite{wang2025svgen}
& - & - & - & - & - & -
& 0.223 & 19.023 & 0.202 & 4.708 \\
OmniSVG-4B~\cite{yang2025omnisvg}
& 0.727 & 0.257 & 0.813 & 19.703 & 0.142 & 4.466
& 0.214 & 19.044 & 0.150 & 4.572 \\
OmniSVG-8B~\cite{yang2025omnisvg}
& 0.764 & 0.229 & 0.853 & 21.401 & 0.172 & 4.541
& 0.229 & 19.101 & 0.153 & 4.662 \\
InternSVG-8B~\cite{wang2025internsvg}
& 0.764 & 0.209 & 0.877 & 22.181 & 0.204 & 4.638
& 0.241 & 19.451 & 0.174 & 4.684 \\
\midrule
\rowcolor{gray!10}
\textbf{HiVG-3B (ours)}
& 0.896 & 0.114 & 0.957 & 21.652 & 0.221 & 4.681
& 0.239 & 20.575 & 0.194 & 4.632 \\
\bottomrule
\end{tabular}
}
\vspace{-1em}
\end{table*}
\subsection{Human Evaluation.}
\noindent Automatic metrics mainly measure raster-domain similarity, but do not fully capture human preference or the practical usability of generated SVG code. We therefore conduct human evaluation from two perspectives: pairwise visual preference and SVG code usability review. 

We randomly sample 60 images from the image-to-SVG test set, covering simple icons, medium-complexity graphics, and more challenging logo- or interface-style compositions, and collect SVG outputs from HiVG-3B and representative open- and closed-source baselines. All results are rasterized at the same resolution for comparison. We recruit 8 professional SVG practitioners as evaluators, and fully randomize method names and output order. In pairwise visual preference, evaluators are shown the reference image and two rendered SVG results, and asked which better reconstructs the reference, with a \emph{tie} allowed. We compare HiVG-3B against SVGen-7B, OmniSVG-8B, InternSVG-8B, Qwen3.5 Plus, Gemini-2.5-pro, GPT-5.2, and Claude-Sonnet-4.5. Each comparison is annotated by 3 evaluators, and the final result is determined by majority vote. Recalling Fig.~\ref{fig:teaser} (c), HiVG achieves the best human evaluation results in
usability and pairwise comparisons against other methods.

\subsection{Ablation Study}
\label{sec:exp_ablation}
\begin{figure*}[t!]
\centering
\includegraphics[width=\textwidth]{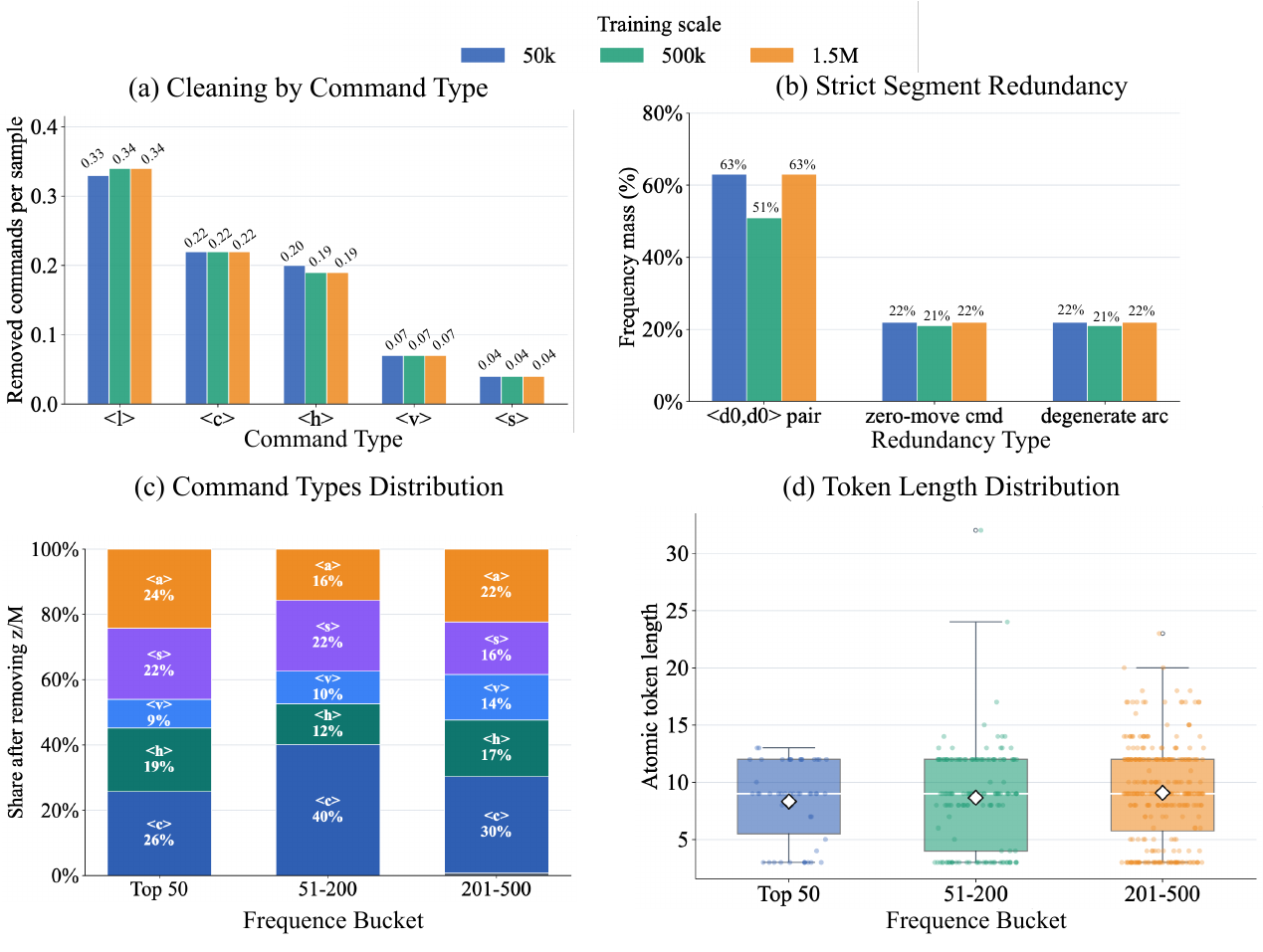}
\vspace{-2em}
\caption{
\textbf{Analysis of structural noise and learned segment properties.} 
(a) \textit{Cleaning statistics:} Commands removed per sample across data scales; noise is primarily concentrated in \texttt{l}, \texttt{c}, and \texttt{h} types. 
(b) \textit{Redundancy patterns:} Frequency mass of strictly degenerate patterns, where \texttt{<d\_0><d\_0>} pairs represent the most prevalent redundant structure. 
(c) \textit{Command distribution:} Relative share of command types within segments across frequency buckets; complex commands like cubic Béziers (\texttt{c}) and arcs (\texttt{a}) are well-captured. 
(d) \textit{Token length:} Atomic token length distribution shows a stable median of $\sim$9 tokens, indicating consistent segment complexity regardless of frequency.
}\label{fig:bpe_clean_tok}
\vspace{-1em}
\end{figure*}

\begin{table*}[h]
\caption{\textbf{Impact of structured SVG modeling.} $\uparrow$ higher is better, $\downarrow$ lower is better. $^\dagger$AR baseline trains on raw SVG string. ``Aes'' denotes the Aesthetic score.}
\vspace{-1em}
\setlength{\tabcolsep}{5pt}
\renewcommand{\arraystretch}{1.15}
\small
\resizebox{\textwidth}{!}{
\begin{tabular}{lcccccccccccc}
\toprule
\multirow{2.5}{*}{Variant} &
\multicolumn{5}{c}{\textbf{Text-to-SVG}} &
\multicolumn{6}{c}{\textbf{Image-to-SVG}} \\
\cmidrule(lr){2-6}\cmidrule(lr){7-12}
& \multicolumn{1}{c}{CLIP$\uparrow$}
& \multicolumn{1}{c}{DINO$\uparrow$}
& \multicolumn{1}{c}{HPS$\uparrow$}
& \multicolumn{1}{c}{PickS$\uparrow$}
& \multicolumn{1}{c}{Aes$\uparrow$}
& \multicolumn{1}{c}{SSIM$\uparrow$}
& \multicolumn{1}{c}{LPIPS$\downarrow$}
& \multicolumn{1}{c}{CLIP-S$\uparrow$}
& \multicolumn{1}{c}{HPS$\uparrow$}
& \multicolumn{1}{c}{PickS$\uparrow$}
& \multicolumn{1}{c}{Aes$\uparrow$} \\
\midrule
AR baseline$^\dagger$ &
0.2146 & 0.1520 & 0.162 & 19.628 & 4.548 & 0.301 & 0.396 & 0.797 & 0.179 & 19.793 & 4.553 \\
\textbf{Ours} &
\textbf{0.2392} & \textbf{0.2795} & \textbf{0.194} & \textbf{20.576} & \textbf{4.632} & \textbf{0.896} & \textbf{0.114} & \textbf{0.957} & \textbf{0.221} & \textbf{21.652} & \textbf{4.681} \\
Improvement &
{\color{green!70!black}+11.5\%} & {\color{green!70!black}+83.9\%} & {\color{green!70!black}+19.8\%} & {\color{green!70!black}+4.8\%} & {\color{green!70!black}+1.8\%} & {\color{green!70!black}+197.7\%} & {\color{green!70!black}-39.3\%} & {\color{green!70!black}+20.1\%} & {\color{green!70!black}+23.5\%} & {\color{green!70!black}+9.4\%} & {\color{green!70!black}+2.8\%} \\
\bottomrule
\end{tabular}
} \label{tab:ablation_svg_modeling}
\vspace{-1em}
\end{table*}
We conduct controlled ablations to verify that each component of HiVG contributes to the final performance under both text-to-SVG and image-to-SVG.
We start by comparing the full structured modeling pipeline against an autoregressive baseline trained on raw SVG strings (Tab.~\ref{tab:ablation_svg_modeling}), establishing the overall gain from modeling SVG as executable programs.
We then probe key design choices: the corpus scale used for Structure Segment Learning (Tab.~\ref{tab:ablation_ssl_scale}), the embedding initialization strategy for new SVG tokens (Tab.~\ref{tab:init_ablation}), and the three-stage curriculum training (Tabs.~\ref{tab:curriculum_image2svg},~\ref{tab:curriculum_text2svg}).
Finally, we analyze whether scaling segment learning introduces undesirable redundancy patterns in learned segments (Fig.~\ref{fig:bpe_clean_tok}).

Overall, these studies reveal three consistent observations: (1) representing SVGs as structured executable programs substantially improves geometric fidelity and semantic alignment; (2) incorporating geometric priors into token embeddings stabilizes early-stage optimization and improves spatial consistency; and (3) progressively increasing sequence complexity through curriculum learning enhances generalization to longer SVG programs.

\noindent\textbf{A. Impact of Structured SVG Modeling.}\quad
To isolate the effect of our proposed structured, geometry-aware SVG modeling pipeline, we compare against an autoregressive baseline trained on raw SVG sequences with a conventional tokenization scheme.
The baseline directly predicts flattened SVG strings with a generic tokenizer, without explicit atomic/segment decomposition.
This ablation assesses whether modeling SVG as a structured executable program yields consistent improvements in semantic alignment, perceptual similarity, and human-preference-related metrics across both generation settings.

\begin{table*}[!t]
\caption{\textbf{Effect of Structure Segment Learning (SSL) corpus scale.}
Dataset statistics and evaluation metrics are reported on both
Text-to-SVG and Image-to-SVG tasks. All models use $M{=}500$ merges.
AT: Atomic Token, ST: Segment Token. Delta rows show changes from previous scale.}
\vspace{-1em}
\setlength{\tabcolsep}{4.5pt}
\renewcommand{\arraystretch}{1.15}
\resizebox{\textwidth}{!}{
\begin{tabular}{l ccc ccc cc cccc cc}
\toprule
\multirow{2.5}{*}{Scale}
& \multicolumn{3}{c}{\textbf{Tokenization Stats}}
& \multicolumn{5}{c}{\textbf{Text-to-SVG}}
& \multicolumn{6}{c}{\textbf{Image-to-SVG}} \\
\cmidrule(lr){2-4}\cmidrule(lr){5-9}\cmidrule(lr){10-15}
& Avg Toks & Raw$\to$AT & AT$\to$ST
& CLIP$\uparrow$ & DINO$\uparrow$
& HPS$\uparrow$ & PickS$\uparrow$ & Aes$\uparrow$
& SSIM$\uparrow$ & LPIPS$\downarrow$ & CLIP-S$\uparrow$
& HPS$\uparrow$ & PickS$\uparrow$ & Aes$\uparrow$ \\
\midrule
$\mathcal{D}_{50\text{k}}$ &
317 & 2.59x & 1.03x &
0.2158 & 0.4072 & 0.157 & 19.398 & 4.412 &
0.696 & 0.313 & 0.803 & 0.174 & 19.710 & 4.398 \\
\midrule
\rowcolor{gray!10}
\textit{$\Delta_{50\text{k}\to500\text{k}}$} &
\textit{+301} & \textit{+0.04 $\times$} & \textit{+0.01 $\times$} &
{\color{green!70!black}\textit{+4.6\%}} & {\color{red!70!black}\textit{-3.3\%}} & {\color{green!70!black}\textit{+9.6\%}} & {\color{green!70!black}\textit{+3.0\%}} & {\color{green!70!black}\textit{+2.4\%}} &
{\color{green!70!black}\textit{+9.1\%}} & {\color{green!70!black}\textit{-26.5\%}} & {\color{green!70!black}\textit{+10.7\%}} & {\color{green!70!black}\textit{+13.8\%}} & {\color{green!70!black}\textit{+5.5\%}} & {\color{green!70!black}\textit{+3.5\%}} \\
\midrule
$\mathcal{D}_{500\text{k}}$ &
618 & 2.63x & 1.04x &
0.2257 & 0.3938 & 0.172 & 19.981 & 4.518 &
0.759 & 0.230 & 0.889 & 0.198 & 20.791 & 4.551 \\
\midrule
\rowcolor{gray!10}
\textit{$\Delta_{500\text{k}\to1.5\text{M}}$} &
\textit{-66} & \textit{0.00 $\times$} & \textit{+0.01 $\times$} &
{\color{green!70!black}\textit{+1.2\%}} & {\color{red!70!black}\textit{-2.3\%}} & {\color{green!70!black}\textit{+3.5\%}} & {\color{green!70!black}\textit{+0.7\%}} & {\color{green!70!black}\textit{+1.0\%}} &
{\color{green!70!black}\textit{+2.4\%}} & {\color{green!70!black}\textit{-9.6\%}} & {\color{green!70!black}\textit{+2.4\%}} & {\color{green!70!black}\textit{+3.5\%}} & {\color{green!70!black}\textit{+1.3\%}} & {\color{green!70!black}\textit{+0.8\%}} \\
\midrule
$\mathcal{D}_{1.5\text{M}}$ &
552 & 2.63x & 1.05x &
\textbf{0.2283} & 0.3848 & \textbf{0.178} & \textbf{20.113} & \textbf{4.564} &
\textbf{0.777} & \textbf{0.208} & \textbf{0.910} & \textbf{0.205} & \textbf{21.056} & \textbf{4.587} \\
\bottomrule
\end{tabular}
} \label{tab:ablation_ssl_scale}
\vspace{-1em}
\end{table*}
\begin{table}[!t]
\caption{\textbf{Ablation on token embedding initialization strategies.}
We report image-to-SVG reconstruction metrics and text-to-SVG generation quality after 1 epoch of training. \textbf{Bold}: best result; \underline{underline}: second best. All methods use the same ${\sim}3{,}000$ SVG tokens and identical training hyperparameters. $^\dagger$Lerp: linear interpolation between text embeddings of ``0'' and ``784'', which does \emph{not} preserve numeric semantics due to character-level BPE tokenization of digit strings.}
\vspace{-1em}
\resizebox{\linewidth}{!}{%
\begin{tabular}{@{}l l cc ccc cccc@{}}
\toprule
\multirow{2}{*}{\#} & \multirow{2}{*}{Method} &
\multicolumn{2}{c}{Components} &
\multicolumn{3}{c}{Img2SVG} &
\multicolumn{4}{c}{Text2SVG} \\
\cmidrule(lr){3-4} \cmidrule(lr){5-7} \cmidrule(lr){8-11}
& & Semantic & Numeric
& LPIPS$\downarrow$ & SSIM$\uparrow$ & CLIP-S$\uparrow$
& CLIP$\uparrow$ & PickScore$\uparrow$ & HPS$\uparrow$ & Aes$\uparrow$ \\
\midrule
1 & \textsc{Noise}        & \xmark & \xmark & \underline{0.226} & 0.440 & 0.795 & 0.207 & 19.831 & 0.144 & 4.250 \\
2 & \textsc{Mean}         & \xmark & \xmark & 0.242 & 0.244 & 0.755 & 0.195 & \textbf{20.110} & \underline{0.142} & \underline{4.830} \\
3 & \textsc{Mean+Noise}   & \xmark & \xmark & 0.237 & \underline{0.523} & \underline{0.821} & 0.205 & 19.645 & 0.137 & 4.785 \\
\midrule
4 & \textsc{Semantic}         & \cmark & \xmark & 0.236 & 0.477 & 0.811 & 0.205 & 19.535 & 0.132 & 4.675 \\
5 & \textsc{Semantic+Noise}   & \cmark & \xmark & 0.233 & 0.550 & 0.830 & \underline{0.206} & 19.585 & 0.135 & 4.715 \\
\midrule
6 & \textsc{HMN} (Lerp)$^\dagger$   & \cmark & \cmark & 0.182 & 0.680 & 0.830 & \underline{0.206} & 19.798 & 0.136 & 4.755 \\
7 & \textsc{HMN} (J-L)    & \cmark & \cmark & \textbf{0.170} & \textbf{0.720} & \textbf{0.880} & \textbf{0.208} & \underline{19.965} & \textbf{0.146} & \textbf{4.870} \\
\bottomrule
\end{tabular}%
}
\label{tab:init_ablation}
\vspace{-1em}
\end{table}

\noindent\textbf{B. Effect of Structure Segment Learning (SSL) Scale.}\quad
\label{sec:ablation_segment_tok_scale}
We investigate how the corpus scale used for Structure Segment Learning affects downstream SVG generation.
Specifically, we learn structure segment merges from three SVG corpora of increasing sizes: 50k, 500k, and 1.5M samples.
For each scale, the resulting segment tokenizer is applied to construct Segment Tokens from Atomic Tokens for both model training and inference, while keeping all other components unchanged.
This study examines whether larger corpora enable more reliable discovery of frequent and structurally valid path segments, leading to improved token composition efficiency, sequence compactness, and geometric fidelity.

\noindent\textbf{C. Effect of Token Initialization Strategy.}\quad
To isolate the contribution of each component in Eq.~\ref{eq:hmn}, we design seven initialization variants with increasing structural priors, summarized in Table~\ref{tab:init_ablation}. All experiments use identical training configurations: Qwen2.5-VL-3B as the backbone, full-parameter SFT with frozen vision encoder and projector, learning rate $10^{-5}$ with the same warmup setting, and 1 epoch on a mixed dataset of image-to-SVG, image-to-caption, and text-to-SVG tasks (${\sim}3{,}000$ domain-specific tokens).
\begin{table*}[!t]
\caption{
\textbf{Effect of three-stage curriculum training on Image-to-SVG with stage-to-stage changes.}
$\uparrow$ higher is better, $\downarrow$ lower is better.  Delta rows show changes from previous stage (positive values indicate improvement for $\uparrow$ metrics).
}
\vspace{-1em}
\resizebox{\textwidth}{!}{
\begin{tabular}{l|cccc|ccc|cccc}
\toprule
& \multicolumn{4}{c|}{\textbf{Validity / Efficiency}}
& \multicolumn{3}{c|}{\textbf{Visual Similarity}}
& \multicolumn{4}{c}{\textbf{Preference / Aesthetic}}
\\
\cmidrule(r){2-5} \cmidrule(r){6-8} \cmidrule(r){9-12}
Method
& Render$\uparrow$
& TokCnt
& PathCnt
& CmdCnt
& SSIM$\uparrow$
& LPIPS$\downarrow$
& CLIP-S$\uparrow$
& ImgR$\uparrow$
& HPS$\uparrow$
& PickS$\uparrow$
& Aes$\uparrow$
\\
\midrule
Stage1-L1
& 95.20\% & 273 & 3.5 & 54.8 & $0.8052 \pm 0.13$ & $0.1716 \pm 0.09$ & $0.9360 \pm 0.07$ & -0.0764 & 0.2110 & 21.348 & 4.6159 \\
Stage1-L2
& 94.20\% & 489 & 5.9 & 107.2 & $0.7223 \pm 0.13$ & $0.2392 \pm 0.09$ & $0.8884 \pm 0.10$ & -0.1836 & 0.2026 & 20.945 & 4.6175 \\
Stage1-L3
& 90.09\% & 656 & 6.7 & 144.2 & $0.7015 \pm 0.13$ & $0.2405 \pm 0.10$ & $0.8750 \pm 0.11$ & -0.2500 & 0.2050 & 21.015 & 4.6285 \\
\midrule
\rowcolor{gray!10}
\textit{$\Delta$S1$\rightarrow$S2}
& \textit{{\color{green!70!black}+0.2\%}} & \textit{+27.0\%} & \textit{+15.3\%} & \textit{+19.5\%} & \textit{{\color{green!70!black}+4.1\%}} & \textit{{\color{green!70!black}-19.7\%}} & \textit{{\color{green!70!black}+4.9\%}} & \textit{{\color{green!70!black}+151.7\%}} & \textit{{\color{green!70!black}+5.9\%}} & \textit{{\color{green!70!black}+21.0\%}} & \textit{{\color{green!70!black}+1.3\%}} \\
\midrule
Stage2-L1
& 93.50\% & 335 & 4.8 & 70.4 & $0.8152 \pm 0.12$ & $0.1611 \pm 0.08$ & $0.9540 \pm 0.06$ & 0.0294 & 0.2150 & 21.522 & 4.6226 \\
Stage2-L2
& 94.40\% & 621 & 6.8 & 128.1 & $0.7520 \pm 0.12$ & $0.1920 \pm 0.08$ & $0.9320 \pm 0.08$ & 0.0950 & 0.2145 & 21.385 & 4.6780 \\
Stage2-L3
& 90.29\% & 897 & 9.6 & 181.5 & $0.7180 \pm 0.13$ & $0.2185 \pm 0.09$ & $0.9210 \pm 0.09$ & 0.0520 & 0.2105 & 21.185 & 4.7150 \\
\midrule
\rowcolor{gray!10}
\textit{$\Delta$S2$\rightarrow$S3}
& \textit{{\color{red}-2.9\%}} & \textit{+26.3\%} & \textit{+16.7\%} & \textit{+27.9\%} & \textit{{\color{red}-2.2\%}} & \textit{{\color{green!70!black}+7.7\%}} & \textit{{\color{green!70!black}+2.7\%}} & \textit{{\color{green!70!black}+186.5\%}} & \textit{{\color{green!70!black}+1.5\%}} & \textit{{\color{green!70!black}+8.4\%}} & \textit{{\color{green!70!black}+0.7\%}} \\
\midrule
Stage3-L1
& 94.60\% & 376 & 6.3 & 85.9 & $0.8129 \pm 0.12$ & $0.1611 \pm 0.08$ & $0.9581 \pm 0.06$ & 0.0742 & 0.2166 & 21.606 & 4.6200 \\
Stage3-L2
& 92.40\% & 825 & 9.7 & 179.9 & $0.7352 \pm 0.12$ & $0.2097 \pm 0.08$ & $0.9471 \pm 0.07$ & 0.1603 & 0.2163 & 21.475 & 4.7050 \\
Stage3-L3
& 87.69\% & 1133 & 11.2 & 232.1 & $0.7019 \pm 0.13$ & $0.2353 \pm 0.08$ & $0.9454 \pm 0.07$ & 0.1490 & 0.2136 & 21.362 & 4.7501 \\
\bottomrule
\end{tabular}
} \label{tab:curriculum_image2svg}
\vspace{-1em}
\end{table*}
\begin{table*}[!t]
\caption{
\textbf{Effect of three-stage curriculum training on Text-to-SVG with stage-to-stage changes.}
$\uparrow$ indicates higher is better, $\downarrow$ indicates lower is better.
Delta rows show changes from previous stage (positive values indicate improvement for $\uparrow$ metrics).
}
\vspace{-1em}
\resizebox{\textwidth}{!}{
\begin{tabular}{l|cccc|c|c|cccc}
\toprule
& \multicolumn{4}{c|}{\textbf{Validity / Efficiency}}
& \multicolumn{1}{c|}{\textbf{Semantic}}
& \multicolumn{1}{c|}{\textbf{Diversity}}
& \multicolumn{4}{c}{\textbf{Preference / Aesthetic}} \\
\cmidrule(r){2-5}
\cmidrule(r){6-6}
\cmidrule(r){7-7}
\cmidrule(l){8-11}
Method
& Render$\uparrow$
& TokCnt$\downarrow$
& PathCnt$\downarrow$
& CmdCnt$\downarrow$
& CLIP$\uparrow$
& DINO$\uparrow$
& ImgR$\uparrow$
& HPS$\uparrow$
& PickS$\uparrow$
& Aes$\uparrow$ \\
\midrule
Stage1-L1
& 95.60\% & 288 & 3.1 & 60.9 & 0.2346 & 0.2949 & -0.4789 & 0.1909 & 20.552 & 4.5891 \\
Stage1-L2
& 95.37\% & 403 & 4.3 & 89.7 & 0.2322 & 0.3535 & -0.7787 & 0.1751 & 20.046 & 4.5663 \\
Stage1-L3
& 94.08\% & 446 & 5.3 & 104.9 & 0.2305 & 0.4015 & -0.8520 & 0.1730 & 19.895 & 4.5580 \\
\midrule
\rowcolor{gray!10}
\textit{$\Delta$S1$\rightarrow$S2}
& \textit{{\color{red}-0.8\%}} & \textit{+58.1\%} & \textit{+37.2\%} & \textit{+52.4\%} & \textit{{\color{green!70!black}+0.6\%}} & \textit{{\color{red}-1.3\%}} & \textit{{\color{green!70!black}+19.3\%}} & \textit{{\color{green!70!black}+5.1\%}} & \textit{{\color{green!70!black}+1.2\%}} & \textit{{\color{green!70!black}+0.6\%}} \\
\midrule
Stage2-L1
& 95.45\% & 428 & 4.3 & 87.6 & 0.2356 & 0.2931 & -0.3768 & 0.1953 & 20.643 & 4.6279 \\
Stage2-L2
& 94.62\% & 637 & 5.9 & 136.7 & 0.2335 & 0.3490 & -0.6285 & 0.1840 & 20.285 & 4.5920 \\
Stage2-L3
& 93.69\% & 717 & 7.7 & 160.6 & 0.2320 & 0.3930 & -0.7180 & 0.1785 & 20.115 & 4.5850 \\
\midrule
\rowcolor{gray!10}
\textit{$\Delta$S2$\rightarrow$S3}
& \textit{{\color{red}-3.5\%}} & \textit{+53.3\%} & \textit{+27.3\%} & \textit{+49.4\%} & \textit{{\color{green!70!black}+0.6\%}} & \textit{{\color{red}-1.8\%}} & \textit{{\color{green!70!black}+2.9\%}} & \textit{{\color{red}-0.3\%}} & \textit{{\color{red}-1.3\%}} & \textit{{\color{green!70!black}+0.7\%}} \\
\midrule
Stage3-L1
& 94.53\% & 532 & 5.8 & 112.2 & 0.2356 & 0.3032 & -0.3776 & 0.1954 & 20.672 & 4.6331 \\
Stage3-L2
& 91.78\% & 943 & 8.5 & 202.5 & 0.2345 & 0.3450 & -0.5380 & 0.1865 & 20.345 & 4.6095 \\
Stage3-L3
& 90.41\% & 1099 & 9.8 & 239.9 & 0.2335 & 0.3859 & -0.6975 & 0.1779 & 20.021 & 4.6164 \\
\bottomrule
\end{tabular}
} \label{tab:curriculum_text2svg}
\end{table*}

\noindent\textbf{D. Impact of Three-Stage Curriculum Training.}\quad
To investigate the effect of curriculum learning on SVG generation, we adopt a three-stage training paradigm based on sequence length.
Specifically, the training corpus is partitioned according to SVG token length into three complexity levels:
\textbf{Stage-1} (30$\sim$326 tokens), 
\textbf{Stage-2} (326$\sim$605 tokens), and 
\textbf{Stage-3} (605$\sim$1k tokens).
The model is progressively trained from shorter to longer sequences.
To evaluate generalization across complexity levels, we construct three test subsets corresponding to these ranges, denoted as \textbf{L1}, \textbf{L2}, and \textbf{L3}.
This design enables fine-grained analysis of how curriculum learning affects performance on simple versus complex SVG structures.
This suggests that gradually increasing program depth stabilizes token embedding learning and improves long-range geometric reasoning.

We observe that curriculum training consistently improves performance on longer sequences (L2/L3) without sacrificing accuracy on simpler cases (L1), suggesting that progressive exposure to structural complexity stabilizes optimization and enhances generalization to high-token-length SVG programs.

\noindent\textbf{E. Structural Noise \& Segment Analysis.}\quad
We further analyze the path-level structural noise identified during Structure Segment Learning (SSL) and the geometric properties of the resulting segments. Figure~\ref{fig:bpe_clean_tok} summarizes the cleaning statistics, redundancy patterns, and segment characteristics across various corpus scales.

As shown in Figure~\ref{fig:bpe_clean_tok} (a), the volume of command-level cleaning remains highly stable across scales, totaling approximately 0.86 removed commands per sample. This consistency suggests that these noise patterns are inherent to raw SVG data rather than artifacts of dataset size. Removals are predominantly concentrated in line-related commands (\texttt{<l>} at $\sim$0.34, \texttt{<h>} at $\sim$0.19) and cubic curves (\texttt{<c>} at 0.22). Figure~\ref{fig:bpe_clean_tok} (b) details the frequency mass of strictly redundant motifs discovered within these segments. Notably, consecutive \texttt{<d\_0><d\_0>} pairs constitute the majority of detected redundancies (51\%--63\%), while zero-move commands and degenerate arcs each account for approximately 22\%.

Beyond noise suppression, SSL effectively captures diverse and meaningful geometric primitives. Figure~\ref{fig:bpe_clean_tok} (c) illustrates the command type distribution within learned segments across frequency buckets. Cubic Bézier curves (\texttt{<c>}) are prominent, particularly in the mid-frequency bucket (51--200) where they reach a 40\% share. High-frequency segments (Top 50) exhibit strong representations of arcs (\texttt{<a>}, 24\%) and smooth curves (\texttt{<s>}, 22\%). Furthermore, Figure~\ref{fig:bpe_clean_tok} (d) reveals that the atomic token length of these segments is remarkably consistent. The median length stays robust at approximately 9 tokens across all frequency tiers, demonstrating SSL's capability to identify compact and structurally stable geometric units while filtering redundant path fragments.

\section{Conclusion}
We introduced a hierarchical tokenization framework for scalable SVG generation. 
By redefining the representation unit from character-level fragments to executable geometric segments, the proposed approach aligns token structure with the semantics of vector graphics. 
The hierarchical design reduces sequence length while preserving structural validity, enabling more stable autoregressive modeling. 
Together with structured initialization and scalable training, the framework demonstrates that representation design plays a crucial role in reliable SVG generation.
Our results suggest that improving geometric consistency does not rely solely on increasing model scale. 
Instead, aligning tokenization with executable structure provides a principled foundation for vector graphics modeling. 
Future work may extend this framework to other structured graphical formats and explore integration with differentiable rendering objectives.

\clearpage
\appendix
\renewcommand{\thefigure}{S\arabic{figure}}
\renewcommand{\thetable}{S\arabic{table}}

\renewcommand{\theHsection}{supp.\Alph{section}}
\renewcommand{\theHfigure}{supp.\arabic{figure}}
\renewcommand{\theHtable}{supp.\arabic{table}}

\renewcommand{\thesubsection}{\Alph{section}.\arabic{subsection}}
\renewcommand{\theHsubsection}{supp.\Alph{section}.\arabic{subsection}}

\setcounter{figure}{0}
\setcounter{table}{0}
\setcounter{page}{1}

\makeatletter
\newcommand{\tocsec}[1]{%
    \vspace{0.6em}\noindent%
    \makebox[1.5em][l]{\textbf{\ref{#1}.}}%
    \textbf{\nameref{#1}} \dotfill \textbf{\pageref{#1}}\par%
}
\newcommand{\tocsubsec}[1]{%
    \vspace{0.2em}\noindent\hspace{1.5em}%
    \makebox[2.5em][l]{\ref{#1}.}%
    \nameref{#1} \dotfill \pageref{#1}\par%
}
\makeatother

\begin{center}
\Large\textbf{Supplementary Material}
\end{center}

\vspace{1em}

\begin{center}
    \Large \textbf{Contents}
\end{center}
\vspace{0.5em}
\hrule height 1pt
\vspace{0.5em}

\tocsec{supp:dataset}


\tocsec{supp:extended_results}
    \tocsubsec{supp:more_results}
    \tocsubsec{supp:comparison}

\tocsec{supp:implementation}
    \tocsubsec{supp:hyperparameters}
    \tocsubsec{supp:prompt_template}

\tocsec{supp:structured_analysis}
    \tocsubsec{supp:noise_patterns}

\vspace{0.8em}
\hrule height 1pt


\section{Dataset Construction, Filtering \& Preprocessing}
\label{supp:dataset}

Our training corpus is built by merging three open-source SVG datasets: SVG-Stack~\cite{rodriguez2025starvector} (2,283,875 samples), SVGX-Dataset~\cite{xing2025empowering} (257,086 samples), and MMSVG-Icon~\cite{yang2025omnisvg} (1,159,423 samples).
After cross-source merging and deduplication, the resulting corpus contains 2,445,092 unique SVG samples.
The merged dataset covers a broad range of vector graphic categories, including icons, emojis, logos, interface elements, and other structured graphic designs.

To improve rendering consistency and reduce malformed or non-executable samples, we apply a unified preprocessing pipeline prior to tokenization.
The pipeline consists of three stages: data cleaning, coordinate transformation, and coordinate quantization.

\noindent\textbf{Data cleaning.}
We first parse each SVG and remove unsupported or unsafe elements.
Specifically, non-renderable or undesirable tags such as \texttt{<foreignObject>} are filtered out, while external-content or executable elements, including \texttt{<image>} and \texttt{<script>}, are rejected.
At the root level, we remove redundant SVG attributes and retain only the \texttt{viewBox} as the canonical geometric reference.
We further inline CSS style rules into element attributes, normalize the SVG structure through a pure-Python preprocessing pipeline, remove unnecessary line breaks, convert color specifications into compact hexadecimal form, and repair missing \texttt{fill} values when necessary.

\noindent\textbf{Coordinate transformation.}
After structural cleaning, all SVGs are mapped into a unified geometric space.
We first expand \texttt{<use>} references by inlining reused elements, ensuring that subsequent transformations operate on explicit geometry only.
For selected light-color SVGs, a dark background may be added to improve rendering visibility.
We then bake all \texttt{transform} attributes directly into coordinates, eliminating residual transformation matrices from the final representation.
Finally, we normalize the \texttt{viewBox} by translating its origin to $(0,0)$ and rescaling the canvas to a target resolution of $784 \times 784$.

\noindent\textbf{Coordinate quantization.}
After global scaling, all coordinates are quantized by rounding to integers.
Absolute coordinates are then converted into relative coordinates to better match the sequential geometric representation used by our tokenizer.
As a final compatibility step, we clip out-of-bound subpaths and clamp minor numerical overflow within a tolerance range of $\pm 10$, which improves tokenizer robustness in borderline cases.

Several implementation choices are important in practice.
First, \texttt{transform} baking is performed only after \texttt{<use>} expansion, preventing duplicated geometric transformations.
Second, coordinate quantization is applied after global scaling to minimize unnecessary precision loss.
Third, boundary correction is deferred to the final stage so that the processed SVGs remain compatible with downstream tokenization and decoding.
Samples that still cannot be parsed, normalized, or rendered stably after preprocessing are discarded.

\begin{table*}[h]
\vspace{-0.5em}
\centering
\caption{\textbf{Expert review of SVG code usability in Adobe Illustrator.}
Eight professional SVG practitioners import the generated SVGs into Adobe Illustrator and score their structural usability on a 1--5 Likert scale.
Higher is better for all metrics.}
\vspace{-0.5em}
\resizebox{\textwidth}{!}{%
\begin{tabular}{lcccc}
\toprule
\textbf{Method}
& \textbf{Semantic Layering} $\uparrow$
& \textbf{Editability} $\uparrow$
& \textbf{Redundancy Control} $\uparrow$
& \textbf{Overall Code Usability} $\uparrow$ \\
\midrule
SVGen-7B~\cite{wang2025svgen}         & 2.88 & 2.83 & 2.74 & 2.82 \\
InternSVG-8B~\cite{wang2025internsvg}     & 3.22 & 3.18 & 3.09 & 3.16 \\
Gemini-2.5-pro~\cite{comanici2025gemini}   & 3.39 & 3.34 & 3.23 & 3.32 \\
GPT-5.2~\cite{openai2025gpt5}          & 3.56 & 3.49 & 3.37 & 3.47 \\
\textbf{HiVG-3B} & \textbf{4.11} & \textbf{4.05} & \textbf{3.96} & \textbf{4.06} \\
\bottomrule
\end{tabular}%
}
\label{tab:supp_code_review}
\vspace{-1em}
\end{table*}

\subsection{SVG Code Usability Review}
\label{supp:code_review}

\noindent\textbf{Motivation and Protocol.}
Raster-domain metrics cannot assess whether a generated SVG remains structurally meaningful and editable after being imported into professional vector-graphics software.
We therefore conduct an additional expert review in Adobe Illustrator~\footnote{We use Adobe Illustrator as a representative industry-standard vector graphics editor for assessing practical SVG editability.}.
The same 8 professional SVG practitioners import the generated SVGs and evaluate their structural usability.
Specifically, they examine whether primitives and path groups correspond to coherent visual-semantic parts, whether local components can be selected and edited conveniently, and whether the SVG contains excessive redundant fragments or implausible decomposition.

Each SVG is scored on a 1--5 Likert scale along four dimensions:
\emph{semantic layering}, \emph{editability}, \emph{redundancy control}, and \emph{overall code usability}.
Because this review is substantially more time-consuming than raster-only inspection, we evaluate five representative methods:
SVGen-7B, InternSVG-8B, Gemini-2.5-pro, GPT-5.2, and HiVG-3B.

\noindent\textbf{Results.}
Table~\ref{tab:supp_code_review} reports the Illustrator-based usability review.
HiVG-3B achieves the best scores on all four dimensions, with the clearest gains in semantic layering and editability.
These results suggest that HiVG improves not only rendered reconstruction quality, but also the structural organization of SVG code in a way that better matches human editing workflows.

\noindent\textbf{Summary.}
Together, the two protocols provide a compact but more complete assessment of image-to-SVG reconstruction.
Pairwise comparison measures what human experts prefer, while Illustrator-based review evaluates structural usability beyond automatic metrics.
Across both settings, HiVG-3B shows consistent advantages, indicating that its improvements extend from raster-domain reconstruction to the practical usability of generated SVG code.

\section{Extended Results}
\label{supp:extended_results}

\begin{figure*}[t]
\centering
\includegraphics[width=1.0\textwidth]{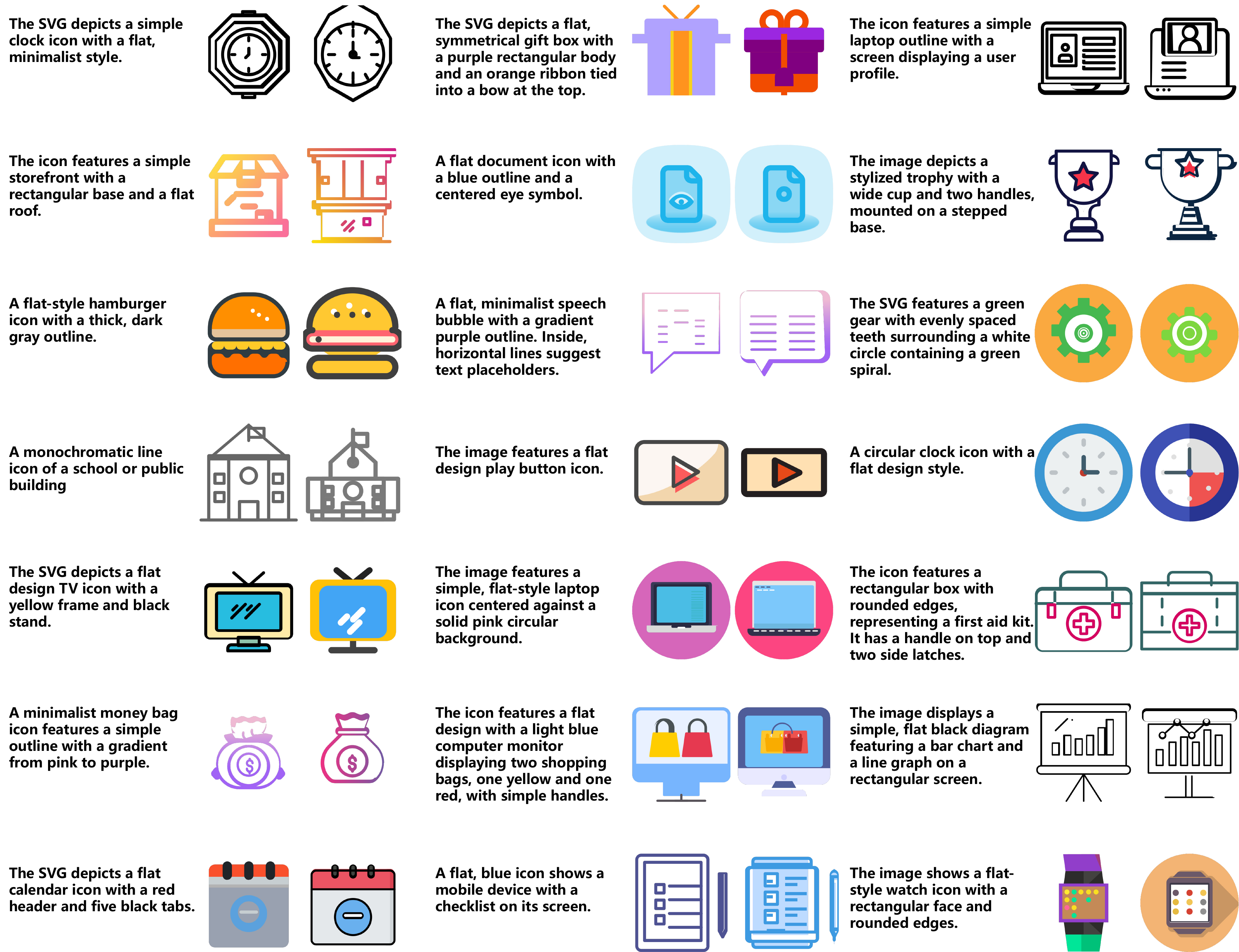}
\vspace{-1em}
\caption{\textbf{Text-to-SVG generation results.}
For each example, we show the text prompt together with the rendered SVG output generated by HiVG.
}
\label{fig:supp_text2svg_comparison}
\vspace{-1em}
\end{figure*}
\begin{figure*}[t]
\centering
\includegraphics[width=1.0\textwidth]{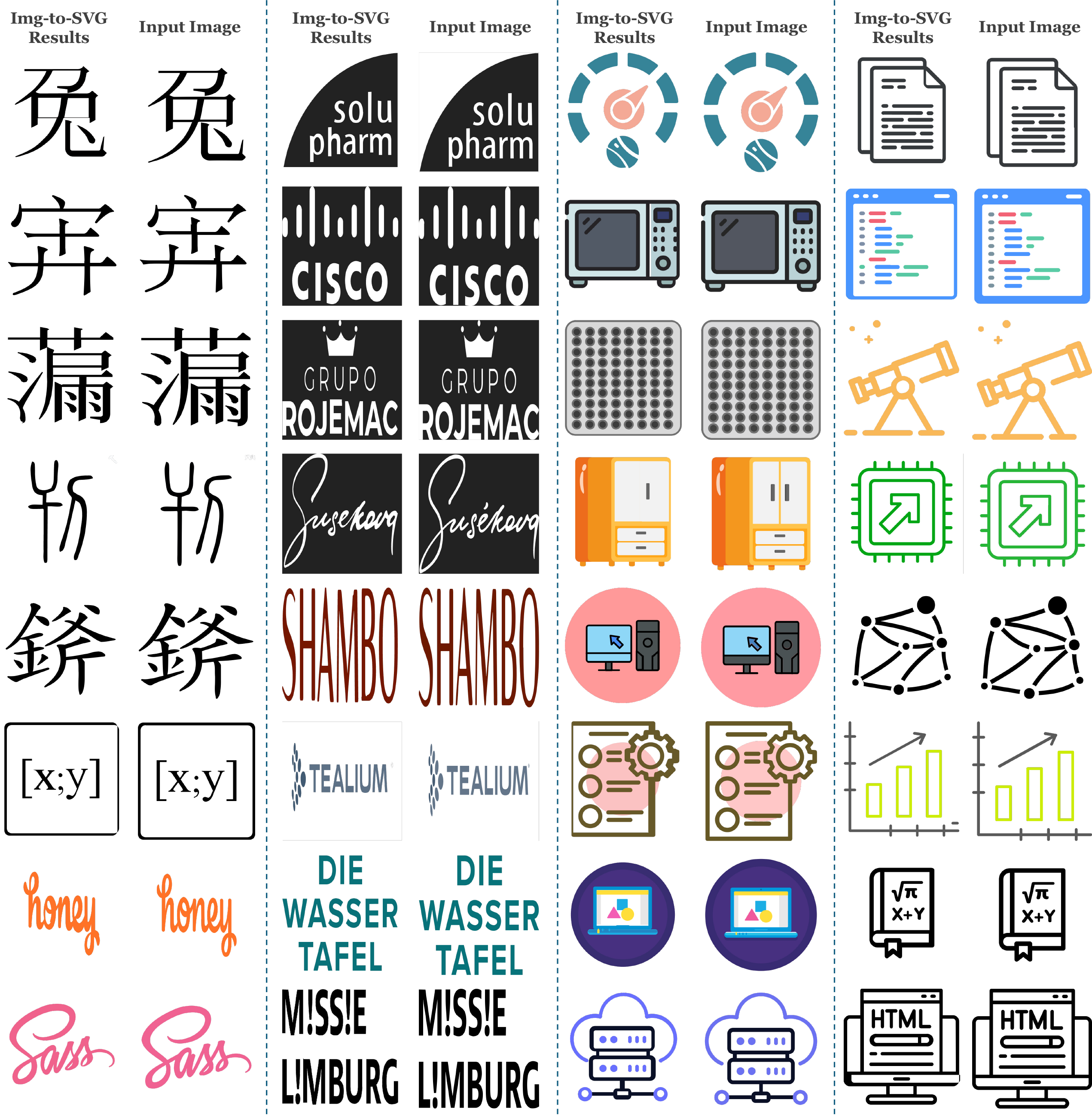}
\vspace{-1em}
\caption{\textbf{Image-to-SVG generation results.}
For each example, the raster input image is shown on the right and the generated SVG rendering on the left.
}
\label{fig:supp_img2svg_comparison}
\vspace{-1em}
\end{figure*}
\begin{figure*}[t]
\centering
\includegraphics[width=1.0\textwidth]{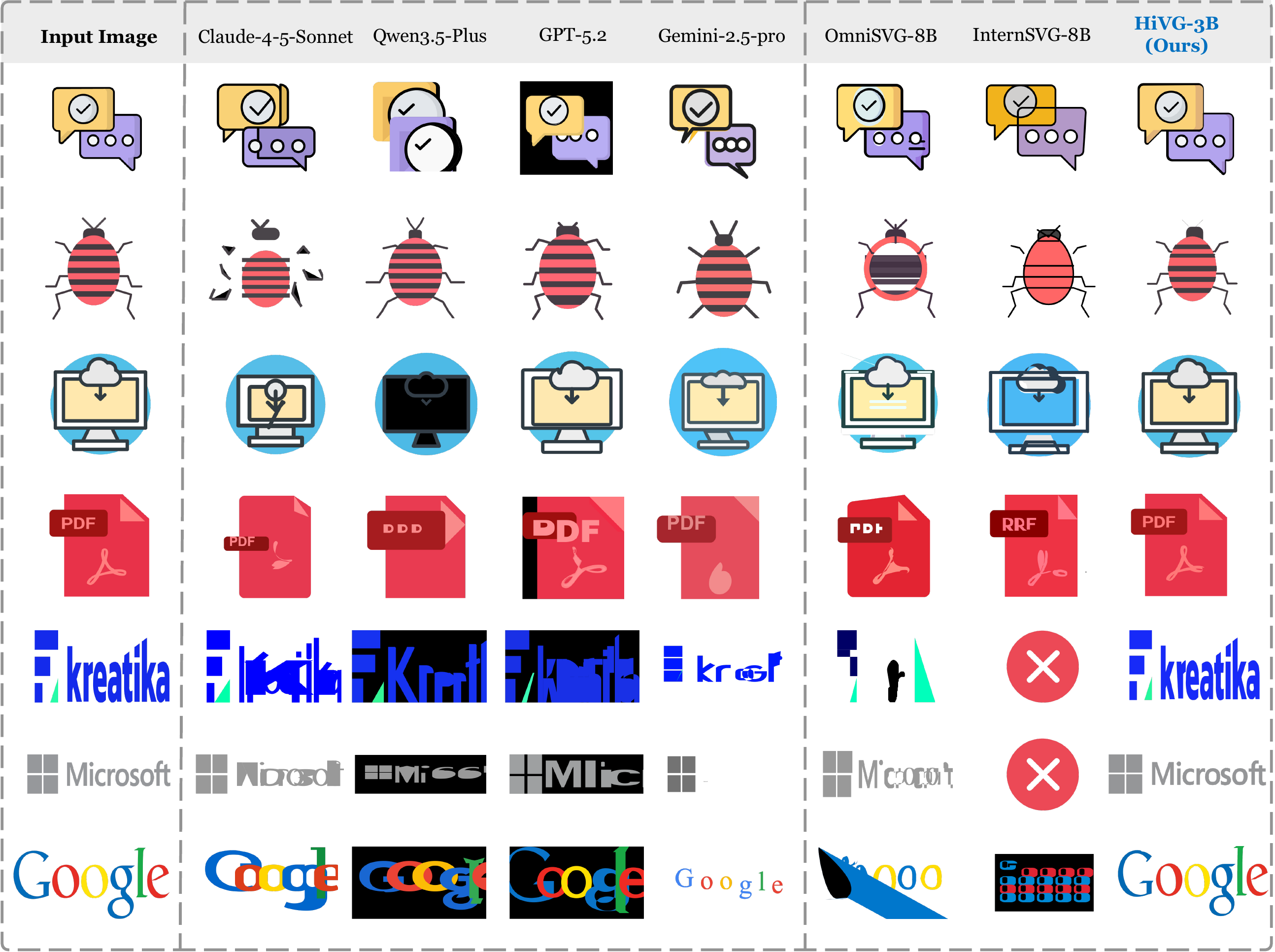}
\vspace{-1.5em}
\caption{\textbf{Additional Image-to-SVG comparison.}
}
\label{fig:more_supp_img2svg_comparison}
\vspace{-1.5em}
\end{figure*}
\begin{figure*}[h!]
\centering
\includegraphics[width=1.0\textwidth]{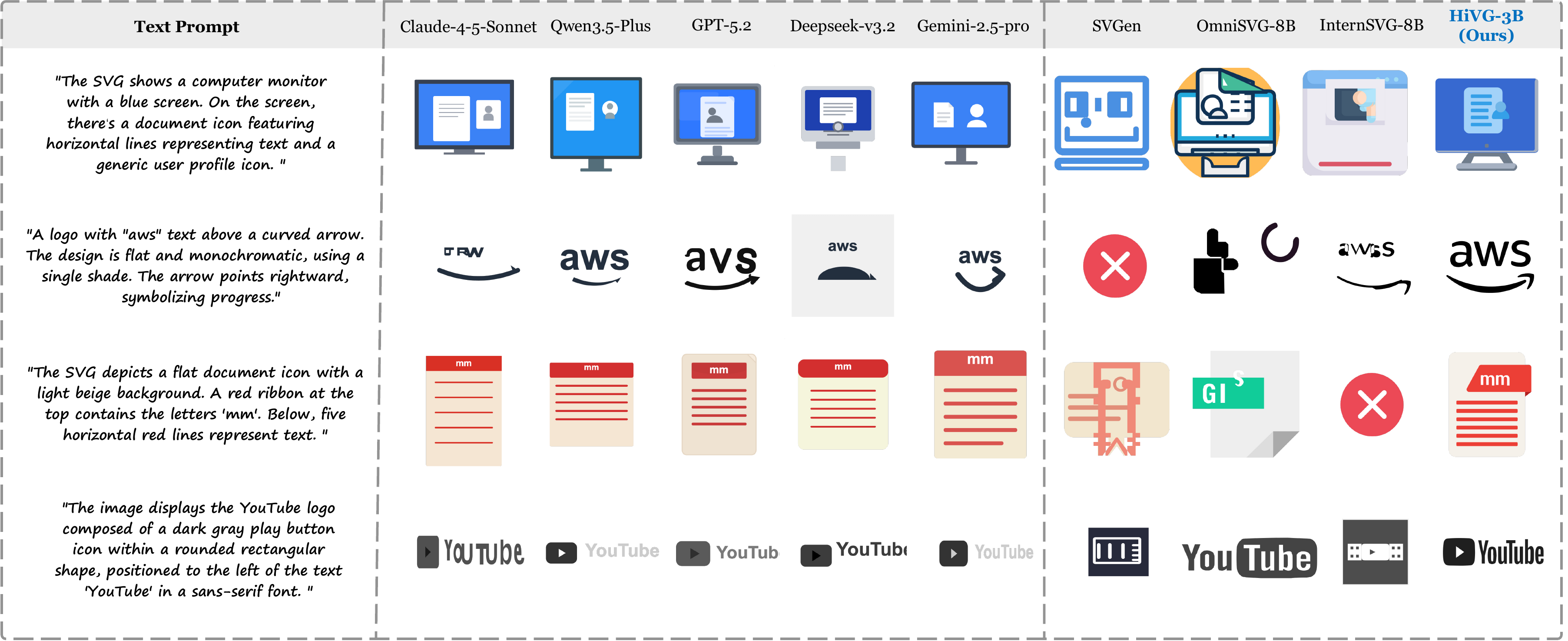}
\vspace{-1.5em}
\caption{\textbf{Additional Text-to-SVG comparison.}
}
\label{fig:more_supp_t2svg_comparison}
\vspace{-1em}
\end{figure*}

\subsection{More Text-to-SVG and Image-to-SVG Results}
\label{supp:more_results}

We provide additional text-to-SVG and image-to-SVG generation examples covering diverse prompts, including flat icons, stylized symbols, logos, and multi-part graphic compositions.
The results in Figure~\ref{fig:ours_text2svg} complement the main paper by illustrating how HiVG handles varying semantic granularity, object composition, and layout structure under open-ended textual descriptions. Compared with the limited examples shown in the main paper, the additional results in Figure~\ref{fig:ours_img2svg} provide a broader view of the model's reconstruction behavior across different levels of geometric complexity.

\subsection{Comparison with Existing Methods}
We provide more text-to-svg and image-to-svg comparison results with existing methods in Figure~\ref{fig:more_supp_img2svg_comparison} \& \ref{fig:more_supp_t2svg_comparison} respectively. 
As visually demonstrated, our method exhibits exceptional proficiency in generating SVGs that contain typographical elements and letters. This success clearly reflects our method's advanced capacity for precise geometric generation and complex topology preservation. Furthermore, achieving such high-quality typography generation with an exceptionally lightweight 3B-parameter model underscores its remarkable efficiency.
\label{supp:comparison}



\section{Extended Implementation Details}
\label{supp:implementation}

\subsection{Initialization Details \& Hyperparameters}
\label{supp:hyperparameters}

The main paper introduces Hierarchical Mean-Noise (HMN) initialization to stably incorporate newly introduced structured SVG tokens into the pretrained language model vocabulary.
Here we provide additional implementation details together with the main hyperparameter settings used in training and inference.
Table~\ref{tab:supp_hparams} summarizes the overall configuration, while the discussion below focuses on the initialization design.

\begin{table}[t]
\centering
\caption{\textbf{Hyperparameter settings of HiVG.}}
\label{tab:supp_hparams}
\vspace{-0.5em}
\resizebox{0.8\textwidth}{!}{%
\begin{tabular}{p{0.56\linewidth}p{0.34\linewidth}}
\toprule
\textbf{Hyperparameter} & \textbf{HiVG} \\
\midrule
\multicolumn{2}{l}{\textit{Architecture / Tokenization}} \\
Backbone model & Qwen2.5-VL-3B-Instruct \\
Canvas size & $784 \times 784$ \\
Atomic token range & $30$--$1000$ \\
Number of curriculum stages & 3 \\
Context length scaling & progressive across stages \\
Atomic vocabulary size & 2450 \\
Segment vocabulary size & 500 \\
Coordinate quantization bins & -794 $\sim$ 794 \\
\midrule
\multicolumn{2}{l}{\textit{Optimization}} \\
Optimizer & AdamW \\
Learning rate & 1e-5 \\
Weight decay & 0.2 \\
Warmup ratio & 0.1 \\
Global batch size & 128 \\
Training epochs & 2 \\
Max context length (S1 / S2 / S3) & 1792 / 2176 / 2432 \\
\midrule
\multicolumn{2}{l}{\textit{HMN Initialization}} \\
Mean anchor weight $\lambda_{\mu}$ & 0.8 \\
Noise scale $\lambda_{n}$ & 0.02 \\
Semantic prior weight $w_{\mathrm{sem}}$ & 0.1 \\
Numeric prior weight $w_{\mathrm{num}}$ & 0.08 \\
Number of RBF bases $K$ & 16 \\
Numeric projection matrix & fixed random \\
\midrule
\multicolumn{2}{l}{\textit{Inference}} \\
Decoding strategy & autoregressive \\
Temperature & 0.7 \\
Top-$p$ & 0.9 \\
Top-$K$ & 50 \\
Repetition penalty & 1.0 \\
Evaluation rendering resolution & 512 $\times$ 512 \\
\bottomrule
\end{tabular}%
}
\vspace{-1em}
\end{table}

In our implementation, the weighting coefficients are set to $\lambda_\mu=[0.8]$, $\lambda_n=[0.02]$, $w_{\mathrm{sem}}=[0.1]$, and $w_{\mathrm{num}}=[0.08]$.
For the numeric projection branch, each normalized scalar value $v_t \in [0,1]$ is expanded using $K=[16]$ Gaussian radial basis functions together with low-order polynomial features, and the resulting vector is projected to the model embedding dimension using a fixed random projection matrix.
This design improves local continuity among coordinate tokens and stabilizes early-stage optimization when the model begins to learn structured SVG geometry.

\lstdefinestyle{chatformat}{
    basicstyle=\small\ttfamily,
    breaklines=true,
    columns=flexible,
    keepspaces=true,
    showstringspaces=false,
    moredelim=[is][\color{teal!70!black}]{<IMAGE>}{</IMAGE>}
}
\begin{figure}[h]
\centering
\setlength{\fboxsep}{0pt}
\begin{tcolorbox}[
    colback=gray!4,
    colframe=gray!70!black,
    boxrule=0.8pt,
    arc=2pt,
    left=2mm,
    right=2mm,
    top=2mm,
    bottom=2mm,
    boxsep=0pt,
    title=\textbf{Text-to-SVG Token Generation (T2ST)},
    fonttitle=\bfseries,
    coltitle=white,
    colbacktitle=gray!70!black,
    toptitle=0.8mm,
    bottomtitle=0.8mm,
    lefttitle=2mm,
    righttitle=2mm,
    width=\textwidth
]
\vspace{-0.5em}
\begin{lstlisting}[style=chatformat,basicstyle=\ttfamily\small,aboveskip=0pt,belowskip=0pt]
<|im_start|>user
[T2ST] Generate SVG tokens from text. {text description}
<|im_end|>
<|im_start|>assistant
<svg>viewBox=0 0 784 784<cmd_M><P_242><P_674><cmd_c><d_7><d_-195>
...</path></svg>
<|im_end|>
\end{lstlisting}
\vspace{-0.5em}
\end{tcolorbox}
\begin{tcolorbox}[
    colback=gray!4,
    colframe=gray!70!black,
    boxrule=0.8pt,
    arc=2pt,
    left=2mm,
    right=2mm,
    top=2mm,
    bottom=2mm,
    boxsep=0pt,
    title=\textbf{Image-to-SVG Token Generation (I2ST)},
    fonttitle=\bfseries,
    coltitle=white,
    colbacktitle=gray!70!black,
    toptitle=0.8mm,
    bottomtitle=0.8mm,
    lefttitle=2mm,
    righttitle=2mm,
    width=\textwidth
]
\vspace{-0.5em}
\begin{lstlisting}[style=chatformat,basicstyle=\ttfamily\small,aboveskip=0pt,belowskip=0pt]
<|im_start|>user
[I2ST] <IMAGE><image></IMAGE>. Generate SVG tokens from image.
<|im_end|>
<|im_start|>assistant
<svg>viewBox=0 0 784 784<cmd_M><P_283><P_567><cmd_c><d_0><d_0>
...</path></svg>
<|im_end|>
\end{lstlisting}
\vspace{-0.5em}
\end{tcolorbox}
\vspace{-1em}
\caption{\textbf{Training templates for text- and image-conditioned SVG token generation.}
For text-to-SVG Token generation (T2ST), the model takes a textual description as input and autoregressively predicts hierarchical SVG tokens.
For image-to-SVG Token generation (I2ST), an image placeholder (\texttt{<image>}) is inserted into the user turn, and the model outputs the corresponding SVG token sequence under the same conversational format.
Using unified templates across both tasks simplifies multi-task training and keeps the supervision interface consistent.}
\label{fig:training_templates}
\vspace{-1em}
\end{figure}
\subsection{Training and Inference Prompt Templates}
\label{supp:prompt_template}
We use unified instruction-style prompts for all training and evaluation settings.
For text-to-SVG generation, the model is prompted to produce SVG code directly from a textual description:

\begin{figure}[h]
\centering
\includegraphics[width=1.0\textwidth]{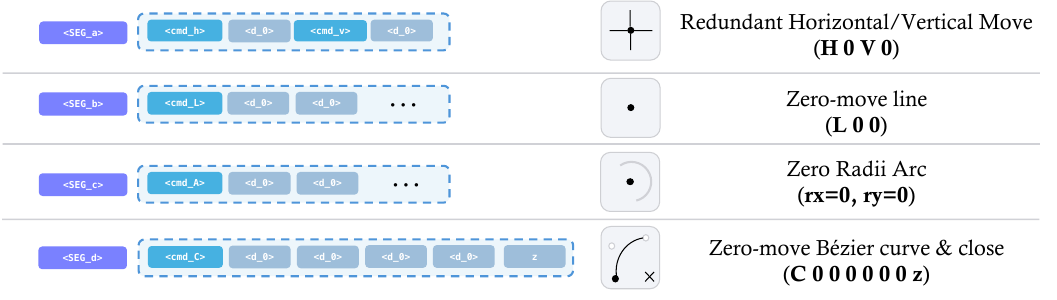}
\vspace{-2em}
\caption{
\textbf{Structural noise patterns in raw SVGs.}
Redundant command groups and zero-move operations contribute no meaningful geometry to the rendered image, yet they artificially inflate token length and degrade computational efficiency.
}
\label{fig:path_noise_example}
\vspace{-1em}
\end{figure}

For image-to-SVG reconstruction, the image token is prepended to the same instruction template, and the model is asked to reconstruct a valid SVG program that faithfully matches the input image.
During evaluation, we use fixed prompt templates across all methods whenever possible, together with unified rendering and post-processing rules, to reduce prompt-induced variance in downstream comparisons.

\section{Additional Analysis of Structured Tokens}
\label{supp:structured_analysis}

\subsection{Path-Level Structural Noise Patterns}
\label{supp:noise_patterns}

To elucidate the learning mechanism of SSL on large-scale SVG corpora, we first investigate the structural noise inherent in raw paths prior to tokenization. Real-world SVGs typically suffer from redundant commands, near-degenerate fragments, and geometric fragmentation, largely artifacts of various authoring tools and conversion pipelines. Although this noise vanishes during rasterization and remains visually unnoticeable, it severely inflates the token sequence length and disrupts the extraction of consistent, reusable segments.

As illustrated in Figure~\ref{fig:path_noise_example}, our cleaning pipeline isolates these irregularities into specific recurring motifs, such as zero-move command groups and redundant transitions that offer no geometric value. Identifying these patterns directly motivates the design of SSL: instead of performing naive text-level compression, our tokenizer is designed to process meaningful executable geometric units, thereby filtering out unstable, non-structural path fragments.

\clearpage
\bibliographystyle{splncs04}
\bibliography{main}

@String(NeurIPS = {Adv. Neural Inform. Process. Syst.})

@String(ICML  = {Int. Conf. Mach. Learn.})

@String(AAAI  = {AAAI})

@String(TMLR  = {Trans. Mach. Learn Res.})

@String(TOG   = {ACM Trans. Graph.})

@String(NeurIPS = {NeurIPS})

@String(ICML  = {ICML})

@String(TMLR  = {TMLR})

@String(TOG   = {ACM TOG})

@article{yang2025omnisvg,
  title={Omnisvg: A unified scalable vector graphics generation model},
  author={Yang, Yiying and Cheng, Wei and Chen, Sijin and Zeng, Xianfang and Yin, Fukun and Zhang, Jiaxu and Wang, Liao and Yu, Gang and Ma, Xingjun and Jiang, Yu-Gang},
  journal={arXiv preprint arXiv:2504.06263},
  year={2025}
}

@inproceedings{xing2025empowering,
  title={Empowering llms to understand and generate complex vector graphics},
  author={Xing, Ximing and Hu, Juncheng and Liang, Guotao and Zhang, Jing and Xu, Dong and Yu, Qian},
  booktitle={Proceedings of the Computer Vision and Pattern Recognition Conference},
  pages={19487--19497},
  year={2025}
}

@article{wang2025internsvg,
  title={Internsvg: Towards unified svg tasks with multimodal large language models},
  author={Wang, Haomin and Yin, Jinhui and Wei, Qi and Zeng, Wenguang and Gu, Lixin and Ye, Shenglong and Gao, Zhangwei and Wang, Yaohui and Zhang, Yanting and Li, Yuanqi and others},
  journal={arXiv preprint arXiv:2510.11341},
  year={2025}
}

@inproceedings{wang2025svgen,
  title={Svgen: Interpretable vector graphics generation with large language models},
  author={Wang, Feiyu and Zhao, Zhiyuan and Liu, Yuandong and Zhang, Da and Gao, Junyu and Sun, Hao and Li, Xuelong},
  booktitle={Proceedings of the 33rd ACM International Conference on Multimedia},
  pages={9608--9617},
  year={2025}
}

@article{xing2025reason,
  title={Reason-SVG: Hybrid Reward RL for Aha-Moments in Vector Graphics Generation},
  author={Xing, Ximing and Guan, Yandong and Zhang, Jing and Xu, Dong and Yu, Qian},
  journal={arXiv preprint arXiv:2505.24499},
  year={2025}
}

@inproceedings{rodriguez2025starvector,
  title={Starvector: Generating scalable vector graphics code from images and text},
  author={Rodriguez, Juan A and Puri, Abhay and Agarwal, Shubham and Laradji, Issam H and Rodriguez, Pau and Rajeswar, Sai and Vazquez, David and Pal, Christopher and Pedersoli, Marco},
  booktitle={Proceedings of the Computer Vision and Pattern Recognition Conference},
  pages={16175--16186},
  year={2025}
}

@article{rodriguez2025rendering,
  title={Rendering-aware reinforcement learning for vector graphics generation},
  author={Rodriguez, Juan A and Zhang, Haotian and Puri, Abhay and Feizi, Aarash and Pramanik, Rishav and Wichmann, Pascal and Mondal, Arnab and Samsami, Mohammad Reza and Awal, Rabiul and Taslakian, Perouz and others},
  journal={arXiv preprint arXiv:2505.20793},
  year={2025}
}

@inproceedings{li2025unisvg,
  title={Unisvg: A unified dataset for vector graphic understanding and generation with multimodal large language models},
  author={Li, Jinke and Yu, Jiarui and Wei, Chenxing and Dong, Hande and Lin, Qiang and Yang, Liangjing and Wang, Zhicai and Hao, Yanbin},
  booktitle={Proceedings of the 33rd ACM International Conference on Multimedia},
  pages={13156--13163},
  year={2025}
}

@article{carlier2020deepsvg,
  title={Deepsvg: A hierarchical generative network for vector graphics animation},
  author={Carlier, Alexandre and Danelljan, Martin and Alahi, Alexandre and Timofte, Radu},
  journal={Advances in Neural Information Processing Systems},
  volume={33},
  pages={16351--16361},
  year={2020}
}

@inproceedings{strokenuwa_tang_2024,
author = {Tang, Zecheng and Wu, Chenfei and Zhang, Zekai and Ni, Minheng and Yin, Shengming and Liu, Yu and Yang, Zhengyuan and Wang, Lijuan and Liu, Zicheng and Li, Juntao and Duan, Nan},
title = {StrokeNUWA: tokenizing strokes for vector graphic synthesis},
year = {2024},
publisher = {JMLR.org},
booktitle = {Proceedings of the 41st International Conference on Machine Learning},
articleno = {1951},
numpages = {16},
location = {Vienna, Austria},
series = {ICML'24}
}

@inproceedings{liu2025freemesh,
title={FreeMesh: Boosting Mesh Generation with Coordinates Merging},
author={Jian Liu and Haohan Weng and Biwen Lei and Xianghui Yang and Zibo Zhao and Zhuo Chen and Song Guo and Tao Han and Chunchao Guo},
booktitle={Forty-second International Conference on Machine Learning},
year={2025},
url={https://openreview.net/forum?id=LBE7HKLQDa}
}

@article{pertsch2025fast,
  title={Fast: Efficient action tokenization for vision-language-action models},
  author={Pertsch, Karl and Stachowicz, Kyle and Ichter, Brian and Driess, Danny and Nair, Suraj and Vuong, Quan and Mees, Oier and Finn, Chelsea and Levine, Sergey},
  journal={arXiv preprint arXiv:2501.09747},
  year={2025}
}

@inproceedings{huang2024compression,
title={Compression Represents Intelligence Linearly},
author={Yuzhen Huang and Jinghan Zhang and Zifei Shan and Junxian He},
booktitle={First Conference on Language Modeling},
year={2024},
url={https://openreview.net/forum?id=SHMj84U5SH}
}

@article{deletang2023language,
  title={Language modeling is compression},
  author={Del{\'e}tang, Gr{\'e}goire and Ruoss, Anian and Duquenne, Paul-Ambroise and Catt, Elliot and Genewein, Tim and Mattern, Christopher and Grau-Moya, Jordi and Wenliang, Li Kevin and Aitchison, Matthew and Orseau, Laurent and others},
  journal={arXiv preprint arXiv:2309.10668},
  year={2023}
}

@inproceedings{huang2024opera,
  title={Opera: Alleviating hallucination in multi-modal large language models via over-trust penalty and retrospection-allocation},
  author={Huang, Qidong and Dong, Xiaoyi and Zhang, Pan and Wang, Bin and He, Conghui and Wang, Jiaqi and Lin, Dahua and Zhang, Weiming and Yu, Nenghai},
  booktitle={Proceedings of the IEEE/CVF Conference on Computer Vision and Pattern Recognition},
  pages={13418--13427},
  year={2024}
}

@inproceedings{bpe_sennrich_2016,
  title={Neural machine translation of rare words with subword units},
  author={Sennrich, Rico and Haddow, Barry and Birch, Alexandra},
  booktitle={Proceedings of the 54th annual meeting of the association for computational linguistics (volume 1: long papers)},
  pages={1715--1725},
  year={2016}
}

@inproceedings{kudo2018sentencepiece,
  title={SentencePiece: A simple and language independent subword tokenizer and detokenizer for neural text processing},
  author={Kudo, Taku and Richardson, John},
  booktitle={Proceedings of the 2018 conference on empirical methods in natural language processing: System demonstrations},
  pages={66--71},
  year={2018}
}

@inproceedings{wang2025cadgpt,
  title={CAD-GPT: Synthesising CAD construction sequence with spatial reasoning-enhanced multimodal LLMs},
  author={Wang, Siyu and Chen, Cailian and Le, Xinyi and Xu, Qimin and Xu, Lei and Zhang, Yanzhou and Yang, Jie},
  booktitle={Proceedings of the AAAI Conference on Artificial Intelligence},
  volume={39},
  number={8},
  pages={7880--7888},
  year={2025}
}

@article{li2020differentiable,
  title={Differentiable vector graphics rasterization for editing and learning},
  author={Li, Tzu-Mao and Luk{\'a}{\v{c}}, Michal and Gharbi, Micha{\"e}l and Ragan-Kelley, Jonathan},
  journal={ACM Transactions on Graphics (TOG)},
  volume={39},
  number={6},
  pages={1--15},
  year={2020},
  publisher={ACM New York, NY, USA}
}

@inproceedings{evolution_tian_2022,
  title={Modern evolution strategies for creativity: Fitting concrete images and abstract concepts},
  author={Tian, Yingtao and Ha, David},
  booktitle={Artificial Intelligence in Music, Sound, Art and Design},
  pages={275--291},
  year={2022},
  organization={Springer}
}

@article{xing2024svgfusion,
  title={Svgfusion: Scalable text-to-svg generation via vector space diffusion},
  author={Xing, Ximing and Hu, Juncheng and Zhang, Jing and Xu, Dong and Yu, Qian},
  journal={arXiv preprint arXiv:2412.10437},
  year={2024}
}

@inproceedings{xing2024svgdreamer,
  title={Svgdreamer: Text guided svg generation with diffusion model},
  author={Xing, Ximing and Zhou, Haitao and Wang, Chuang and Zhang, Jing and Xu, Dong and Yu, Qian},
  booktitle={Proceedings of the IEEE/CVF conference on computer vision and pattern recognition},
  pages={4546--4555},
  year={2024}
}

@inproceedings{jain2023vectorfusion,
  title={Vectorfusion: Text-to-svg by abstracting pixel-based diffusion models},
  author={Jain, Ajay and Xie, Amber and Abbeel, Pieter},
  booktitle={Proceedings of the IEEE/CVF Conference on Computer Vision and Pattern Recognition},
  pages={1911--1920},
  year={2023}
}

@inproceedings{frans2022clipdraw,
title={{CLIPD}raw: Exploring Text-to-Drawing Synthesis through Language-Image Encoders},
author={Kevin Frans and Lisa Soros and Olaf Witkowski},
booktitle={Advances in Neural Information Processing Systems (NeurIPS)},
year={2022},
}

@article{clipasso_vinker_2022,
  title={Clipasso: Semantically-aware object sketching},
  author={Vinker, Yael and Pajouheshgar, Ehsan and Bo, Jessica Y and Bachmann, Roman Christian and Bermano, Amit Haim and Cohen-Or, Daniel and Zamir, Amir and Shamir, Ariel},
  journal={ACM Transactions on Graphics (TOG)},
  volume={41},
  number={4},
  pages={1--11},
  year={2022},
  publisher={ACM New York, NY, USA}
}

@article{xing2023diffsketcher,
  title={Diffsketcher: Text guided vector sketch synthesis through latent diffusion models},
  author={Xing, Ximing and Wang, Chuang and Zhou, Haitao and Zhang, Jing and Yu, Qian and Xu, Dong},
  journal={Advances in Neural Information Processing Systems},
  volume={36},
  pages={15869--15889},
  year={2023}
}

@inproceedings{wu2025chat2svg,
  title={Chat2svg: Vector graphics generation with large language models and image diffusion models},
  author={Wu, Ronghuan and Su, Wanchao and Liao, Jing},
  booktitle={Proceedings of the Computer Vision and Pattern Recognition Conference},
  pages={23690--23700},
  year={2025}
}

@article{qwen2.5_2024,
  title={Qwen2. 5 technical report},
  author={Yang, An and Yang, Baosong and Zhang, Beichen and Hui, Binyuan and Zheng, Bo and Yu, Bowen and Li, Chengyuan and Liu, Dayiheng and Huang, Fei and Wei, Haoran and others},
  journal={arXiv preprint arXiv:2412.15115},
  year={2024}
}

@article{qwen2.5vl_2025,
  title={Qwen2. 5-vl technical report},
  author={Bai, Shuai and Chen, Keqin and Liu, Xuejing and Wang, Jialin and Ge, Wenbin and Song, Sibo and Dang, Kai and Wang, Peng and Wang, Shijie and Tang, Jun and others},
  journal={arXiv preprint arXiv:2502.13923},
  year={2025}
}

@misc{qwen3_5_2026,
  title = {{Qwen3.5}: Towards Native Multimodal Agents},
  author = {{Qwen Team}},
  month = {February},
  year = {2026},
  url = {https://qwen.ai/blog?id=qwen3.5}
}

@misc{google2025gemini,
  title={{Gemini} 3 {Pro} Model Card},
  author={{Google DeepMind}},
  year={2025},
  howpublished={\url{https://deepmind.google/models/gemini/pro/}},
  note={{Model Card}, December 2025}
}

@article{comanici2025gemini,
  title={Gemini 2.5: Pushing the frontier with advanced reasoning, multimodality, long context, and next generation agentic capabilities},
  author={Comanici, Gheorghe and Bieber, Eric and Schaekermann, Mike and Pasupat, Ice and Sachdeva, Noveen and Dhillon, Inderjit and Blistein, Marcel and Ram, Ori and Zhang, Dan and Rosen, Evan and others},
  journal={arXiv preprint arXiv:2507.06261},
  year={2025}
}

@misc{openai2025gpt5,
  title={Update to {GPT-5} {System Card}: {GPT-5.2}},
  author={OpenAI},
  year={2025},
  howpublished={\url{https://openai.com/index/introducing-gpt-5-2/}},
  note={{System Card}, December 2025}
}

@misc{claude45sonnet_modelcard_2025,
  title = {Claude 4.5 Model Card},
  author = {{Anthropic}},
  year = {2025},
  url = {https://www-cdn.anthropic.com/claude-4-5-model-card.pdf},
  note = {Accessed: 2026-03-04}
}

@misc{deepseekai2025deepseekv32,
  title = {DeepSeek-V3.2: Pushing the Frontier of Open Large Language Models},
  author = {{DeepSeek-AI}},
  year = {2025},
  eprint = {2512.02556},
  archivePrefix = {arXiv},
  primaryClass = {cs.CL},
  url = {https://arxiv.org/abs/2512.02556}
}

@misc{aesthetic_christoph_2022,
author = {Christoph Schuhmann},
title = {Improved Aesthetic Predictor},
year = {2022},
booktitle = {Github Repo},
howpublished = {\url{https://github.com/christophschuhmann/improved-aesthetic-predictor}}
}

@inproceedings{clip_Radford_2021,
  title={Learning transferable visual models from natural language supervision},
  author={Radford, Alec and Kim, Jong Wook and Hallacy, Chris and Ramesh, Aditya and Goh, Gabriel and Agarwal, Sandhini and Sastry, Girish and Askell, Amanda and Mishkin, Pamela and Clark, Jack and others},
  booktitle={International Conference on Machine Learning},
  pages={8748--8763},
  year={2021},
  organization={PMLR}
}

@article{hpsv2_Wu_2023,
  title={Human preference score v2: A solid benchmark for evaluating human preferences of text-to-image synthesis},
  author={Wu, Xiaoshi and Hao, Yiming and Sun, Keqiang and Chen, Yixiong and Zhu, Feng and Zhao, Rui and Li, Hongsheng},
  journal={arXiv preprint arXiv:2306.09341},
  year={2023}
}

@article{dinov2_oquab_2024,
title={{DINO}v2: Learning Robust Visual Features without Supervision},
author={Maxime Oquab and Timoth{\'e}e Darcet and Th{\'e}o Moutakanni and Huy V. Vo and Marc Szafraniec and Vasil Khalidov and Pierre Fernandez and Daniel HAZIZA and Francisco Massa and Alaaeldin El-Nouby and Mido Assran and Nicolas Ballas and Wojciech Galuba and Russell Howes and Po-Yao Huang and Shang-Wen Li and Ishan Misra and Michael Rabbat and Vasu Sharma and Gabriel Synnaeve and Hu Xu and Herve Jegou and Julien Mairal and Patrick Labatut and Armand Joulin and Piotr Bojanowski},
journal={Transactions on Machine Learning Research (TMLR)},
issn={2835-8856},
year={2024},
url={https://openreview.net/forum?id=a68SUt6zFt},
}

@article{xu2023imagereward,
  title={Imagereward: Learning and evaluating human preferences for text-to-image generation},
  author={Xu, Jiazheng and Liu, Xiao and Wu, Yuchen and Tong, Yuxuan and Li, Qinkai and Ding, Ming and Tang, Jie and Dong, Yuxiao},
  journal={Advances in Neural Information Processing Systems},
  volume={36},
  pages={15903--15935},
  year={2023}
}

@article{kirstain2023pickApic,
  title={Pick-a-pic: An open dataset of user preferences for text-to-image generation},
  author={Kirstain, Yuval and Polyak, Adam and Singer, Uriel and Matiana, Shahbuland and Penna, Joe and Levy, Omer},
  journal={Advances in neural information processing systems},
  volume={36},
  pages={36652--36663},
  year={2023}
}

@article{ghojogh2021johnson,
  title={Johnson-Lindenstrauss lemma, linear and nonlinear random projections, random Fourier features, and random kitchen sinks: Tutorial and survey},
  author={Ghojogh, Benyamin and Ghodsi, Ali and Karray, Fakhri and Crowley, Mark},
  journal={arXiv preprint arXiv:2108.04172},
  year={2021}
}

@inproceedings{randomfeature2007rahimi,
author = {Rahimi, Ali and Recht, Benjamin},
title = {Random features for large-scale kernel machines},
year = {2007},
isbn = {9781605603520},
publisher = {Curran Associates Inc.},
address = {Red Hook, NY, USA},
booktitle = {Proceedings of the 21st International Conference on Neural Information Processing Systems},
pages = {1177–1184},
numpages = {8},
location = {Vancouver, British Columbia, Canada},
series = {NIPS'07}
}

\end{document}